\documentclass[10pt,twocolumn,letterpaper]{article}

\usepackage[review]{cvpr}      

\usepackage[dvipsnames]{xcolor}

\definecolor{cvprblue}{rgb}{0.21,0.49,0.74}
\usepackage[pagebackref,breaklinks,colorlinks,citecolor=cvprblue]{hyperref}

\def\paperID{14044} 
\def\confName{CVPR}
\def\confYear{2025}

\usepackage{adjustbox}
\usepackage{threeparttable}
\usepackage{xcolor}
\usepackage{multirow}
\usepackage{makecell}
\usepackage{siunitx}
\usepackage{tcolorbox}
\usepackage[accsupp]{axessibility} 
\usepackage{microtype}
\usepackage{marvosym}
\usepackage{cuted}
\usepackage{multirow}
\usepackage{makecell}
\usepackage{colortbl}
\usepackage{tabulary}
\usepackage{pifont} 
\usepackage{fvextra}
\usepackage{listings}

\usepackage{CJKutf8}

\usepackage[utf8]{inputenc}
\usepackage[vietnamese,english]{babel}
\usepackage{graphicx}

\addto{\captionsenglish}{
  
}

\newcommand{\redcross}{\textcolor{red}{\ding{55}}}
\newcommand{\greentick}{{\color{green}\ding{51}}}
\newcommand{\ccocr}{{\sc {CC-OCR}}\xspace}

\newcommand\mypara[1]{\vspace{1.0mm}\noindent\textbf{#1}}

\title{\ccocr: A Comprehensive and Challenging OCR Benchmark for Evaluating Large Multimodal Models in Literacy}

\author{Zhibo Yang\textsuperscript{1}\thanks{Equal Contribution.} \quad Jun Tang\textsuperscript{1}\footnotemark[1] \quad Zhaohai Li\textsuperscript{1}\footnotemark[1] \quad Pengfei Wang\textsuperscript{1}\footnotemark[1] \quad Jianqiang Wan\textsuperscript{1}\footnotemark[1] \quad
Humen Zhong\textsuperscript{1}\footnotemark[1] \quad \\ Xuejing Liu\textsuperscript{1} \quad Mingkun Yang\textsuperscript{1} \quad Peng Wang\textsuperscript{1} 
\quad Shuai Bai\textsuperscript{1} 
\quad Lianwen Jin\textsuperscript{2}
\quad Junyang Lin\textsuperscript{1}\thanks{Corresponding Author.} \\
\\
\quad \textsuperscript{1} Alibaba Group 
\quad \textsuperscript{2} South China University of Technology
}

\begin{document}
\maketitle

\begin{strip}\centering
\vspace{-18mm}
\includegraphics[width=1.0\textwidth]{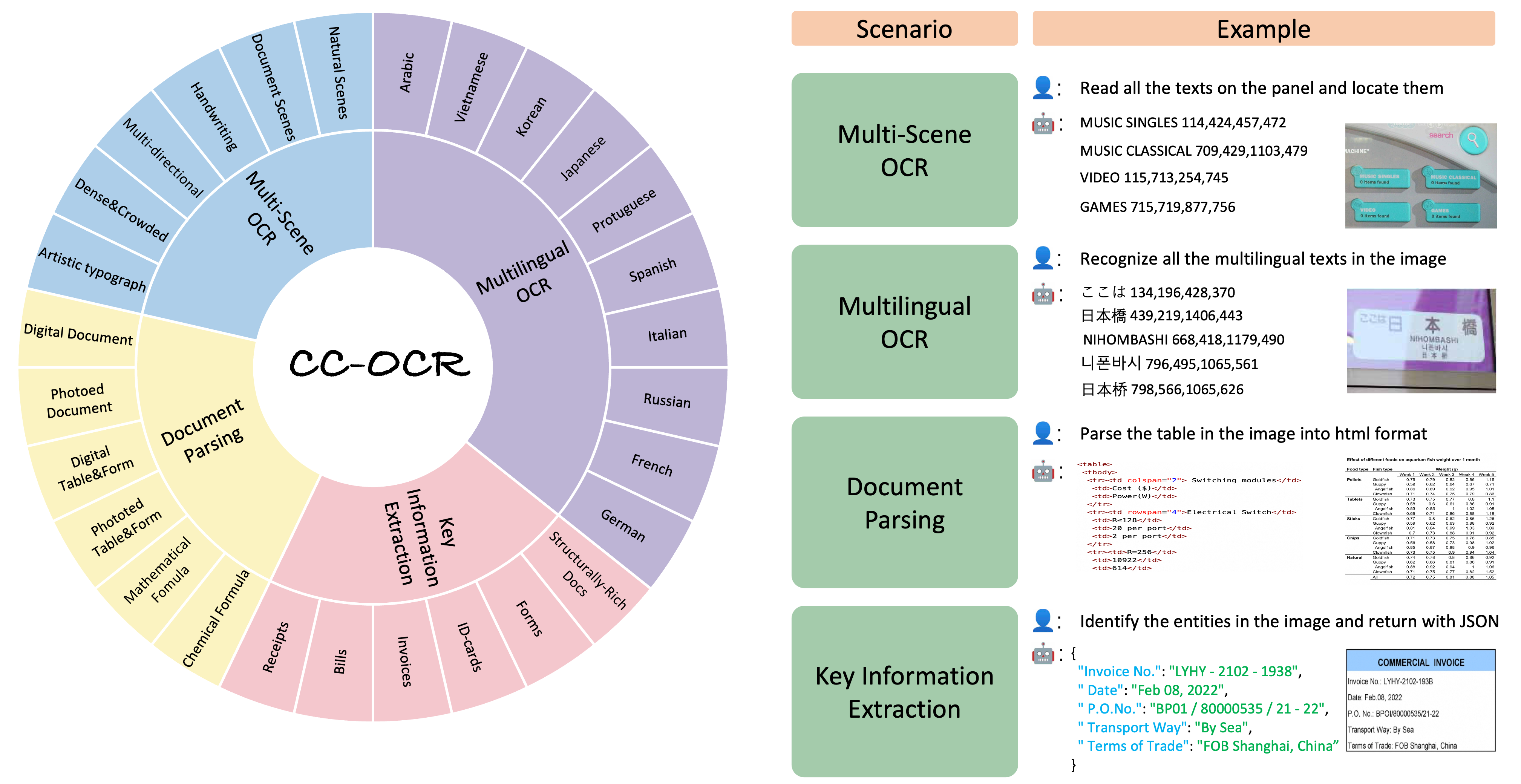}
\captionof{figure}{Task Taxonomy of CC-OCR.
\label{fig:cover_fig}}
\vspace{-2mm}
\end{strip}

\begin{abstract}
Large Multimodal Models ( LMMs ) have demonstrated impressive performance
in recognizing document images with natural language instructions. However, it remains unclear to what extent capabilities in literacy with rich structure and fine-grained visual challenges. The current landscape lacks a comprehensive benchmark to effectively measure the literate capabilities of LMMs. Existing benchmarks are often limited by narrow scenarios and specified tasks. To this end, we introduce \textbf{CC-OCR}, a comprehensive benchmark that possesses a diverse range of scenarios, tasks, and challenges. \textbf{CC-OCR} comprises four OCR-centric tracks: multi-scene text reading, multilingual text reading, document parsing, and key information extraction. It includes 39 subsets with 7,058 full annotated images, of which 41\% are sourced from real applications, and released for the first time. We evaluate nine prominent LMMs and reveal both the strengths and weaknesses of these models, particularly in text grounding, multi-orientation, and hallucination of repetition. \textbf{CC-OCR} aims to comprehensively evaluate the capabilities of LMMs on OCR-centered tasks, facilitating continued progress in this crucial area. 
\vspace{-2mm}

\end{abstract}

\section{Introduction}
\label{sec:intro}
Literate capability is crucial in today's LMMs~\cite{team2023gemini,achiam2023gpt,bai2023qwen,chen2024internvl}, as it empowers models to engage with texts in images and multimodal dialogues critically. 
The ability to literate not only includes reading plain texts but also involves position grounding, structured layouts, and handling formatted input and structured output.
LMMs hold great potential to unify these literate capabilities using natural language instructions and images as prompts, which meanwhile leads to heightened demand for innovative and comprehensive evaluations. In the OCR domain, specialized benchmarks exist to evaluate various downstream tasks such as text line recognition~\cite{iiit5k_2012bmvc_mishra,ic15_2015icdar_karatzas}, text detection~\cite{ch2017total,he2018icpr2018}, formula recognition~\cite{mahdavi2019icdar,yuan2022syntax}, multilingual recognition~\cite{nayef2019icdar2019}, and table structure recognition~\cite{li-etal-2020-tablebank,long2021parsing}. However, these benchmarks are not applicable for evaluating LMMs due to they are fragmented, purpose-diverse, and inconsistently annotated. Recent benchmarks like TextVQA~\cite{singh2019towards} and DocVQA~\cite{mathew2021docvqa} aim to evaluate textual comprehension capabilities. Nevertheless, these benchmarks fall short of thoroughly and quantitatively evaluating their perceptual and interpretative literacy. OCRBench~\cite{liu2023hidden} is a benchmark for evaluating the line-level text recognition capabilities of LMMs, but it does not capture the broader full-image challenges.


\begin{figure*}[th]
    \includegraphics[width=1\linewidth]{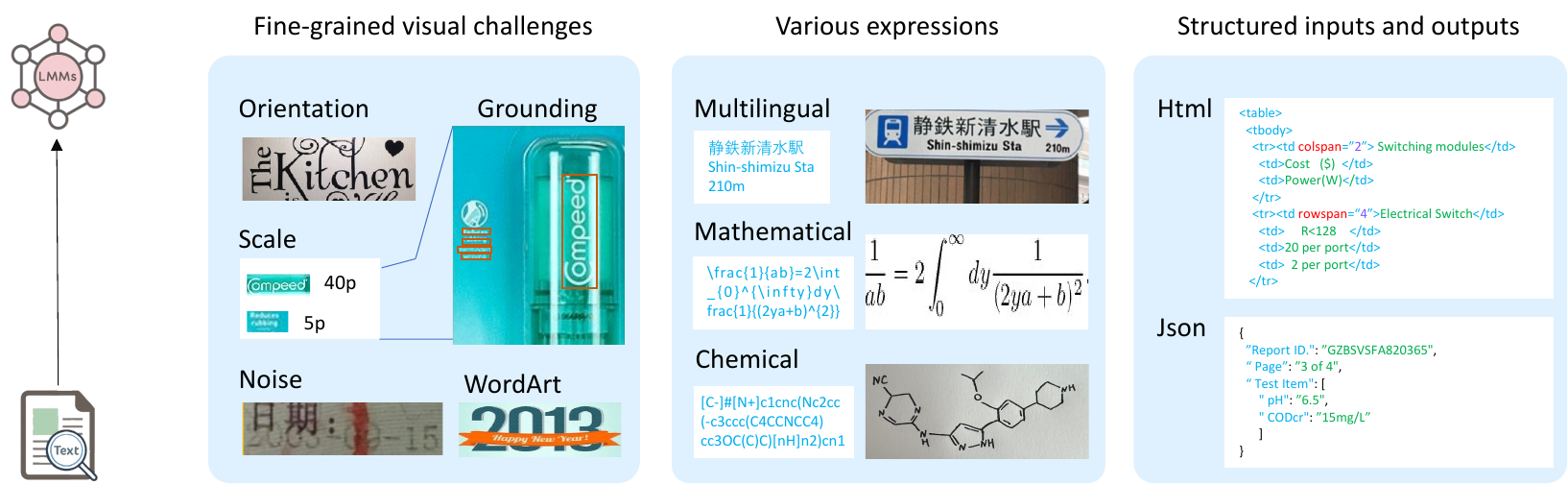}
    \centering
    \caption{Challenge Taxonomy of CC-OCR. 
    }
    \label{fig:challenges}
    \vspace{-3mm}
\end{figure*}

To bridge this critical gap, we introduce CC-OCR, a comprehensive and challenging benchmark designed specifically to evaluate LMMs across different dimensions of literacy capabilities, including multilingual text reading, text spotting, layout-aware document parsing, and key information extraction. These capabilities can pave LMMs with the pathway for real-world
applications, including document digitization, office automation, and urban monitoring.

CC-OCR is distinguished by two key features: the comprehensiveness of tasks and the challenges across various scenarios. 
On the one hand, from an application perspective, as presented in Fig.~\ref{fig:cover_fig}, we focus on four OCR-centric tasks. On the other hand, to highlight challenges, as shown in Fig.~\ref{fig:challenges}, \ccocr covers fine-grained visual challenges ( i.e., orientation-sensitivity, grounding, natural noise, and artistic text), decoding of various expressions, and structured inputs and outputs.

For thorough evaluation, we selected five generalist models, including GPT-4o~\cite{achiam2023gpt}, Gemini-1.5~\cite{team2023gemini}, Claude-3.5-Sonnet~\cite{TheC3}, Qwen2-VL-72B~\cite{wang2024qwen2}, and InternVL2~\cite{chen2024internvl}. We also test four specialist models, including KOSMOS2.5~\cite{lv2023kosmos}, TextMonkey~\cite{liu2024textmonkey}, Florence~\cite{xiao2024florence}, and GOT~\cite{wei2024got}.

Our comprehensive evaluation reveals that although
LMMs achieve remarkable performance on high-level scene understanding
tasks like full-text captioning, current
LMMs decrease to deliver satisfactory performance in challenging text-rich images. Gemini-1.5-Pro and Qwen2-VL-72B are the two top-performing models. Gemini-1.5-Pro achieves first in the multi-scene, multilingual, and parsing tracks, while Qwen2-VL-72B takes first place in KIE and second place in multi-scene OCR and parsing. Our findings also reveal several interesting insights: (1) Text in natural scenes remains underexplored, with performance being over 15\% lower than those for documents. (2) Structured formats are challenging to recognize, as evidenced by the poorer performance in parsing and KIE tasks compared to text-only recognition. (3) Most models exhibit weaker multilingual capabilities compared to their proficiency in Chinese and English. (4) Fine-grained text grounding is a common limitation across all models.

To summarize, our major contributions are as follows:
\begin{itemize}
    \item We introduce CC-OCR, a comprehensive and challenging benchmark tailored to OCR, encompassing four primary tasks and 39 sub-tasks, featuring multi-scene, multilingual, and multi-format output.
    \item We conduct a thorough evaluation of nine different LMMs, including five generalist LMMs and four specialist LMMs designed for document, providing a quantitative analysis of their strengths and weaknesses.    
    \item Our evaluation reveals valuable insights guidance for enhancing LMMs to achieve advanced levels of literate intelligence.
\end{itemize}

\vspace{-2mm}
\section{Related Work}
\label{sec:related}
\vspace{-1.5mm}
\mypara{Multi-Scene Text Reading}. There are a variety of specialized benchmarks in the early era of deep learning, such as natural-scene~\cite{ic15_2015icdar_karatzas,shi2017icdar2017}, web-scene~\cite{he2018icpr2018}, multi-directional and curved text~\cite{yuliang2017detecting}. Current OCRBench~\cite{liu2023hidden} for evaluating LMMs mainly focuses on line-granularity recognition. DocLocal4K~\cite{hu2024mplug} and FOX~\cite{liu2024focus} mainly curate data from document images. KOSMOS-2.5~\cite{lv2023kosmos} proposes two benchmarks: OCREval and MarkdownEval, but both lack data from natural scenes. These benchmarks focus on a specific challenge and therefore fail to serve as comprehensive benchmarks for evaluating general-purpose LMMs. 

\mypara{Multilingual Text Reading} emphasizes the capability of text recognition across various languages. Related multilingual evaluation datasets are scarce, especially those for whole images. MLT 2019~\cite{nayef2019icdar2019}, widely used for multilingual text reading, provides a diverse set of images with text in multiple languages. 
MTVQA~\cite{tang2024mtvqa} integrates visual and textual reasoning through question-answering (QA) pairs, adding complexity to the evaluation. However, it lacks comprehensive OCR annotations, focusing mainly on QA pairs, which limits its utility for OCR tasks. 


\mypara{Document Parsing} encompasses a diverse range of scenarios and tasks, including formula recognition~\cite{blecher2023nougat}, table~\cite{zhao2024tabpedia} and chart analysis~\cite{chen2024onechart}, element detection~\cite{zhao2024doclayout}, and layout analysis~\cite{luo2024layoutllm, tang2023unifying}. Most LMMs evaluate their recognition capabilities exclusively on scanned documents, overlooking real-world documents. The factors in real-shot documents pose significant challenges for text content and structural recognition and limit the robustness of current models. 

\mypara{Key Information Extraction.}
Currently, the KIE data can be divided into two categories, constrained-category and open-category datasets, depending on whether the key to be extracted is predefined. For the constrained KIE, the target categories to be extracted are given. There are public datasets available, such as SROIE~\cite{Huang2019ICDAR2019CO}, CORD, EPHOIE~\cite{Wang2021TowardsRV}, and POIE~\cite{kuang2023visual}. These benchmarks also provide end-to-end representation (i.e. dictionary form) which can be adopted to evaluate the large models. SIBR~\cite{yang2023modeling} and HUST-CELL~\cite{yu2023icdar} are typical open-category benchmarks, the structures of which are represented by entities and various types of entity links. Therefore, this annotation format cannot be directly used to evaluate large models. 

\vspace{-2mm}
\section{Dataset}
\vspace{-1.5mm}
\subsection{Dataset Collection}
\begin{figure*}[th]
    \includegraphics[width=0.8\linewidth]{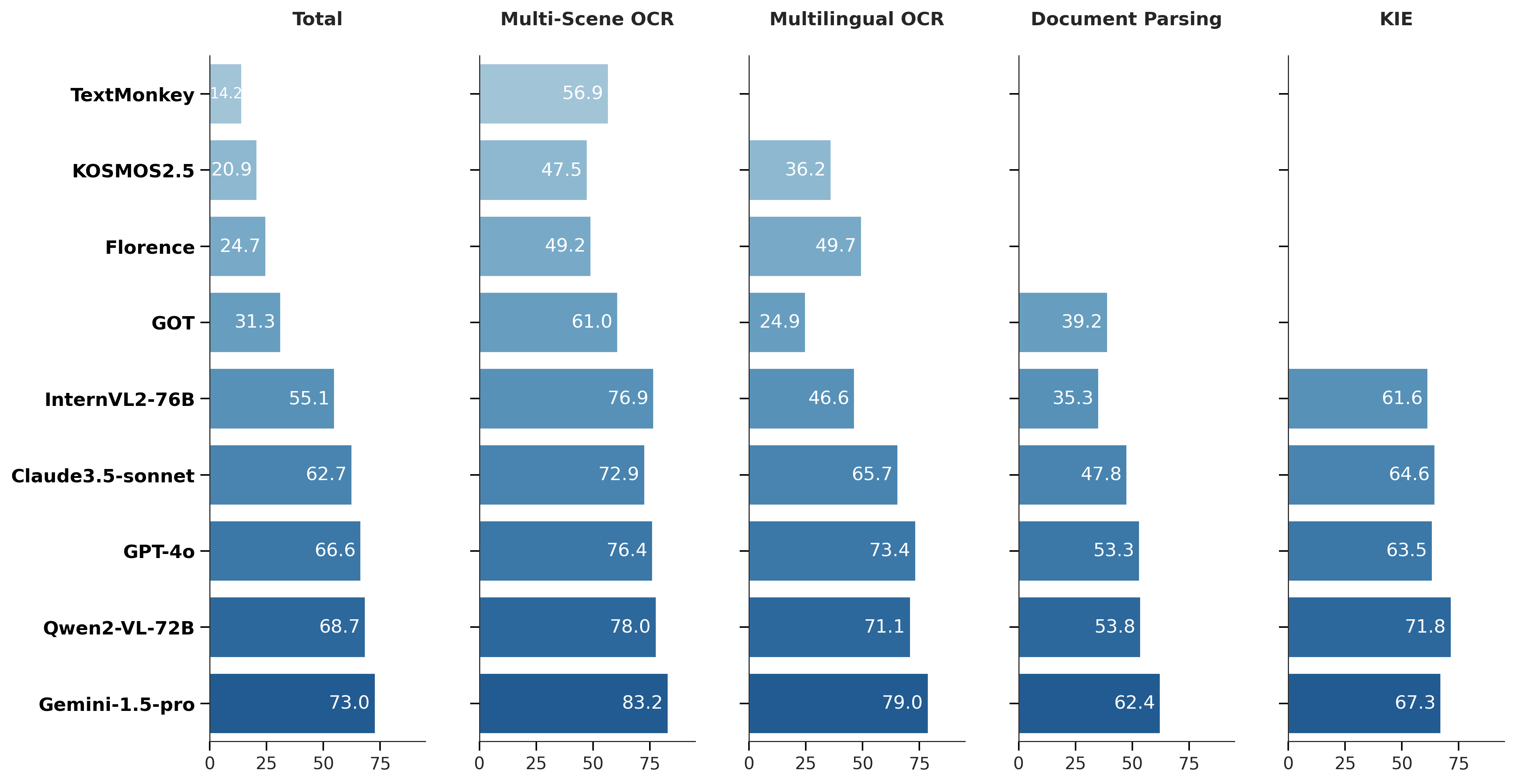}
    \centering
     \vspace{-1em}
    \caption{Comparative Analysis of Large Models on Four Tracks and an Overall Performance. }
    \label{fig:overall_performance}
    \vspace{-1.5em}
\end{figure*}

The CC-OCR dataset is constructed following three criteria: diversity, practicality, and challenge. In alignment with these criteria, we prioritize using natural scene images, real-world captured documents, and handwritten images. The data sources consist of three parts: existing benchmarks with qualified annotations, re-annotated datasets meeting LMM evaluation standards, and newly-collected data. The detailed composition of \ccocr is presented in Appendix.~\ref{sec:details_ccocr}.

\mypara{Multi-Scene OCR} benchmark primarily comprises data in both English and Chinese. The English segment leverages established academic benchmark datasets, which encompass natural scene (TotalText~\cite{ch2017total}, IC15~\cite{ic15_2015icdar_karatzas}, HierAgent), documents (FUNSD~\cite{jaume2019funsd}, CORD~\cite{park2019cord}, IAM~\cite{marti2002iam} ), multi-orientation (InverseText~\cite{ye2023inversetext}), and artistic text (LAION-OCR~\cite{schuhmann2022laion}). In contrast, the Chinese component consists entirely of newly collected datasets, encompassing street scene (Scene-zh), web scene (UGC-ver-zh, DAST-zh~\cite{tang2019dast}), and document (Doc-zh, Doc-hw-zh). HierAgent is a subset of Hiertext~\cite{long2022hiertext}, consisting of data related to mobile devices, computer screen-shots, and machine control interfaces. The LAION-OCR includes 400 randomly selected and re-annotated images from LAION-5B~\cite{schuhmann2022laion}. In total, this sub-benchmark contains 1850 natural scene images, 350 document images, and 550 web images. 

\mypara{Multilingual OCR}. In this sub-benchmark, each language consists of 10 real-shot document images and 140 natural scene images. This ratio is strategically designed to achieve a balance in the number of words. The document images are entirely newly-collected, featuring multi-orientation and real-shot. The natural scene images are partially derived from the re-annotation of the MTVQA~\cite{tang2024mtvqa} and partially sourced from newly-collected data, particularly for Russian, Spanish, and Portuguese.



\mypara{Document Parsing} benchmark includes three types of data: 300 document images, 300 table images, and 200 formula images. All data is sourced from in-house collections. Table and formula datasets refer to those documents where the main content is composed of tables or formulas. In contrast, the dataset of \textit{documents} is primarily made up of plain texts, with only a small portion of the areas containing tables and formulas.  


\mypara{Key Information Extraction} benchmark can be divided into two types: constrained-category with 1008 images and open-category with 1000 images.
For the constrained-category, data are derived from SROIE, CORD, EPHOIE, and POIE. For balance, we sample the 250 samples from POIE. Open-category images are derived from SIBR and HUST-CELL. However, the original annotation of these two datasets designed to evaluate entities and entity relationships is no longer applicable. We re-annotated these two datasets using a standardized JSON format to maintain consistency with the previously constrained-category datasets. The newly annotated open-category dataset is renamed as \textbf{SOLD}, short for \textbf{S}tructurally-rich \textbf{O}pen \textbf{L}ayout \textbf{D}ataset.


\subsection{Dataset Comparison}
The evaluation of Large Multimodal Models (LMMs) for OCR tasks requires comprehensive benchmarks that can assess various capabilities. While numerous OCR datasets emerged during the deep learning era, they were typically designed for specific tasks rather than holistic evaluation. To position our work, we compare CC-OCR with four recent contemporary benchmarks (OCRBench, Fox, DocLocal4K, and KOSMOS2.5-Eval), as detailed in Tab.~\ref{tab:bench_comp}. 

\begin{table*}[ht]    
    \centering
    \small
    \caption{Comparisons of different benchmarks for evaluating LMMs.  }
    \begin{tabular}{ccccccccccc}
        \toprule
         \multirow{2}{*}{Benchmark} & \multirow{2}{*}{Granularity} & \multirow{2}{*}{Language} & \multicolumn{4}{c}{Scene} & \multicolumn{3}{c}{Track}  & \multirow{2}{*}{Num} \\
         \cmidrule(lr){4-7} \cmidrule(lr){8-10}
         & &  & Doc & Natural Scene & Web & Handwritten & OCR & Parsing & KIE & \\ 
         \midrule
        OCRBench~\cite{liu2023hidden} & Line Image & 2 & \greentick & \greentick & \greentick & \greentick & \greentick & \redcross & \greentick & 1,000 \\
        FOX~\cite{liu2024focus} & Full Image & 2 & \greentick & \greentick & \redcross & \redcross & \greentick & \redcross & \redcross & 612 \\
         DocLocal4K~\cite{hu2024mplug} & Full Image & 1 & \greentick & \greentick & \greentick & \redcross & \greentick & \redcross & \redcross & 4,250 \\       
         KOSMOS2.5-Eval~\cite{lv2023kosmos} & Full Image & 1 & \greentick & \redcross & \greentick & \redcross & \greentick & \greentick & \redcross & 7,990 \\        
         \midrule
         CC-OCR & Full Image & 10 & \greentick & \greentick & \greentick & \greentick & \greentick & \greentick & \greentick & 7,058 \\
         \bottomrule
         
    \end{tabular}

    \label{tab:bench_comp}

\end{table*}

\begin{figure*}[h]
    \includegraphics[width=0.8\linewidth]{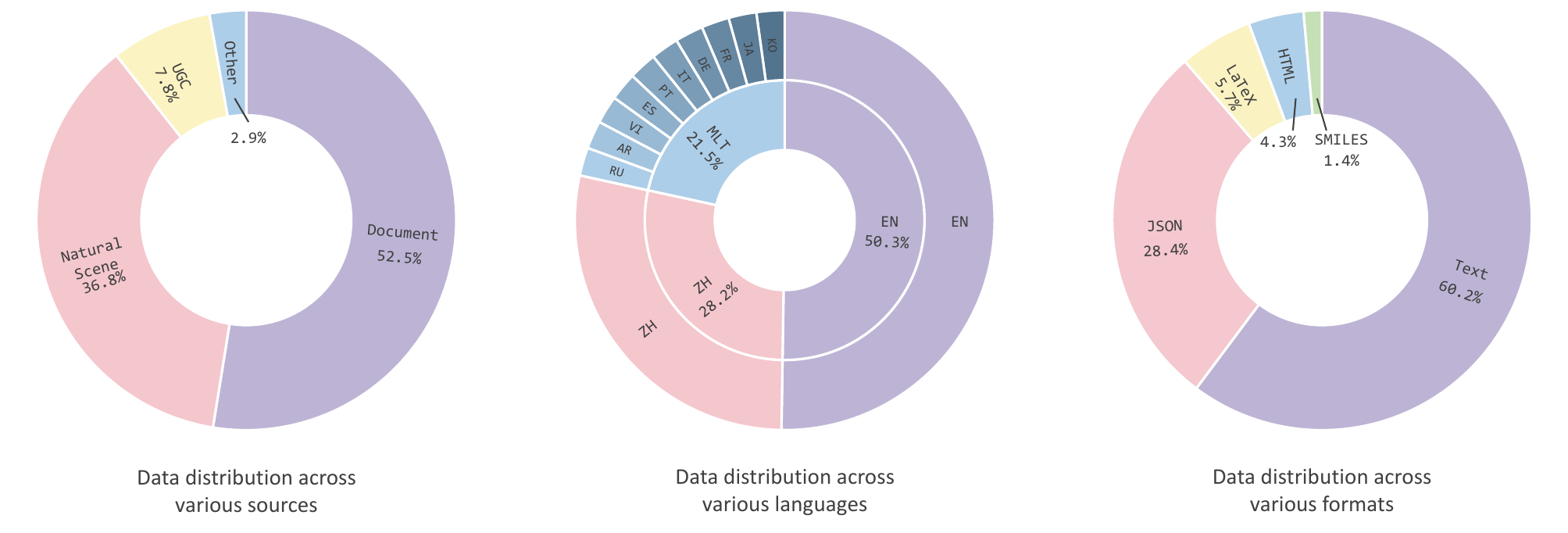}
    \centering
    \caption{Statistics of Data Distribution. 
    }
    \label{fig:data_analysis}
    \vspace{-1.5em}
\end{figure*}

One difference among these benchmarks lies in their text processing granularity. Most existing benchmarks focus on full-image analysis, with OCRBench being the exception in its emphasis on line-level recognition and single-question VQA tasks. This granular approach, while valuable for specific applications, limits the comprehensive evaluation of LMM capabilities.  
Language coverage represents another significant distinguishing factor. DocLocal4K and KOSMOS2.5-Eval are limited to English-only evaluation, while OCRBench and Fox extend their coverage to include both English and Chinese. CC-OCR substantially broadens this linguistic scope by supporting ten major languages spanning multiple language families. These include Indo-European languages (English, German, French, Spanish, Portuguese, Russian), East Asian languages (Chinese, Japanese, Korean), Arabic from the Afroasiatic family, and Vietnamese from the Austroasiatic family.
The benchmarks also differ markedly in their coverage of real-world scenarios. While most existing benchmarks lack support for one or more key scenarios, OCRBench and CC-OCR encompass all four essential scenarios.  
In terms of evaluation capabilities, most benchmarks focus solely on OCR performance. KOSMOS2.5-Eval extends this by including both OCR and parsing tracks. Our CC-OCR further expands the evaluation scope by incorporating three distinct tracks: OCR performance, document parsing, and key information extraction.  
In general, the proposed CC-OCR benchmark distinguishes itself by offering the most comprehensive evaluation framework among existing benchmarks. Its multi-dimensional coverage across languages, scenarios, and evaluation tracks enables a thorough assessment of LMM capabilities in diverse real-world applications.

\subsection{Dataset Annotation}
Based on the categorization of elements and structures present in the images, alongside the intended output objectives, we classify the annotations into four distinct types: texts, structured documents, tables, and information extraction. An iterative annotation refinement process is employed for all new annotations. The process begins with model-based pre-annotation, proceeds with cross-validation by different models, and culminates in manual correction to ensure annotation accuracy.

\mypara{Annotation of Texts}. The text annotations cover multi-scene and multilingual OCR, for which we provide text transcriptions and word-level grounding (line-level bounding boxes for languages such as Chinese and Japanese).

\mypara{Annotation of Documents}. For documents, due to their diverse elements, we use LaTeX format for representation. Firstly, we utilize the Mathpix-Markdown-IT tool to convert image content into the LaTeX format. We retain the plain text, formulas, and tables, while style-related symbols are removed. For handwritten formulas and chemical molecular formulas, we first generate pseudo-labels using a recognition model and then manually correct any recognition errors. The annotations for these data types are formatted in LaTeX code and SMILES format, respectively. During evaluation, the accuracy of tables and formulas within the document sub-set is also assessed using edit distance.

\mypara{Annotation of Structured Tables}. For the evaluation of table structures, we use the HTML format. This is because HTML can structurally represent and evaluate various complex structures, such as row and column spanning. HTML evaluation is applied to all table subsets of document parsing.

\mypara{Annotation of Information Extraction.} As mentioned in the data collection section, all extraction tasks are uniformly represented in JSON format. For the open-vocabulary and open-structured datasets, SOLD-SIBR and SOLD-CELL, we conduct a new round of annotation. We add annotations for hierarchical structures, table-format key-value (KV) groups, and items without keys. Finally, the annotation results are reviewed and verified using both OCR experts and large language models. For a detailed process, please refer to Appendix.~\ref{sec:details_anno_kie}.  

\subsection{Dataset Analysis}
Appendix.~\ref{sec:details_ccocr} presents the detailed composition of the proposed CC-OCR benchmark. The benchmark consists of four main tracks: Conventional OCR, Multilingual OCR, Document Parsing, and Key Information Extraction. All Chinese data in Multi-Scene OCR are newly collected. The multilingual data is composed equally of re-annotated open-source data and newly collected data. All document parsing data is newly collected. For KIE, the data of the constrained-category has been re-annotated.

To better demonstrate the balance of the dataset, we conduct a three-dimensional analysis of its distribution, with the results presented in Fig.~\ref{fig:data_analysis}. In terms of the source scenarios, 52.5\% of the data are derived from documents, evenly divided between digital and real-shot documents. Natural scene data accounts for 36.8\%, while web images make up 7.8\%. From a linguistic perspective, English accounts for the largest proportion at 50.3\%, followed by Chinese at 28.2\%. The remaining 10 languages are evenly distributed, collectively comprising the remaining 21.5\%. Regarding data format, plain text constitutes 60.2\% of the dataset, JSON format accounts for 28.4\%, LaTeX for 5.7\%, HTML for 4.3\%, and 1.4\% is the SMILES format for molecular formula.

\section{Evaluation Method}
For evaluating OCR, both text sequences and positional sequences, we employ evaluation metrics similar to those used in TextMonkey~\cite{liu2024textmonkey}, namely Eval-Trans and Eval-Pos. Given that the majority of large models exhibit limited grounding capabilities, we primarily use Eval-Trans in our main experiments. For document parsing and formula recognition, we employ the Normalized Edit Distance~\cite{lcvenshtcin1966binary} (NED) for evaluation. For table parsing, we use Tree Edit Distance-based Similarity (TEDS)~\cite{zhong2020image}. For Key information extraction, we follow previous work~\cite{kim2022ocr,hwang2019post} and use field-level F1 score for evaluation. Detailed evaluation methods can be found in Appendix.~\ref{sec:details_eval}.

\section{Experiments}

\begin{table*}[ht]    
    \small
    \centering
    \caption{Results of LMMs on subsets of Multi-Scene OCR track. The upper part of the table refers to the specialist models, while the lower part refers to the generalist models.}    
    \begin{tabulary}{\textwidth}{CCCCCCCCCCCCCCC}
        \toprule
         \multirow{2}{*}{Method} & \multicolumn{5}{c}{Natural Scene} & \multicolumn{5}{c}{Document} & \multicolumn{3}{c}{Web UGC} & \multirow{2}{*}{Total} \\
         \cmidrule(lr){2-6} \cmidrule(lr){7-11} \cmidrule(lr){12-14}
            & \scalebox{0.7}[0.7]{ TotalText} & \scalebox{0.7}[0.7]{ IC15} & \scalebox{0.7}[0.7]{ InverseText} & \scalebox{0.7}[0.7]{ HierAgent} & \scalebox{0.7}[0.7]{ Scene\textsubscript{zh}} & \scalebox{0.7}[0.7]{ FUNSD } & \scalebox{0.7}[0.7]{ CORD} & \scalebox{0.7}[0.7]{ IAM} & \scalebox{0.7}[0.7]{ Doc\textsubscript{zh}} & \scalebox{0.7}[0.7]{Doc-hw\textsubscript{zh}} & \scalebox{0.7}[0.7]{ LAION\textsubscript{OCR}} & \scalebox{0.7}[0.7]{ UGC-ver\textsubscript{zh}} & \scalebox{0.7}[0.7]{ DAST\textsubscript{zh}} &  \\ 
         \midrule
         \scalebox{0.8}[0.8]{ KOSMOS2.5~\cite{lv2023kosmos}} & 37.51 & 36.46 & 16.18 & 49.93 & 11.94 & \textbf{90.34} & 90.56 & 90.51 & 59.09 & 13.69 & 66.67 & 13.14 & 42.17 & 47.55 \\          
         \scalebox{0.8}[0.8]{Florence~\cite{xiao2024florence}} & 72.76 & 61.89 & 53.23 & 63.94 & 22.28 & 73.46 & 79.35 & 79.9 & 15.7 & 9.56 & 85.32 & 3.41 & 19.29 & 49.24 \\
         \scalebox{0.8}[0.8]{TextMonkey~\cite{liu2024textmonkey}} & 69.84 & 46.11 & 45.18 & 51.19 & 55.17 & 76.39 & 79.76 & 69.96 & 54.17 & 26.67 & 83.16 & 33.99 & 47.84 & 56.88 \\
         \scalebox{0.8}[0.8]{GOT~\cite{wei2024got}} & 36.2 & 46.1 & 20.94 & 47.07 & 65.23 & 83.87 & 66.98 & 65.33 & 75.56 & 63.17 & 62.7 & 82.34 & 77.57 & 61.00 \\
         \midrule         
         \scalebox{0.8}[0.8]{Claude-3.5-Sonnet~\cite{TheC3}} & 76.09 & 48.1 & 39.67 & 66.64 & 53.62 & 85.73 & 92.32 & \textbf{96.02} & 82.33 & 66.09 & 85.51 & 80.04 & 75.14 & 72.87 \\         
         \scalebox{0.8}[0.8]{GPT-4o\cite{achiam2023gpt}} & 79.82 & 61.52 & 77.03 & 61.85 & 58.3 & 86.08 & 86.46 & 95.23 & 77.96 & 61.72 & \textbf{86.45} & 87.64 & 73.12 & 76.40 \\
         \scalebox{0.8}[0.8]{InternVL2-76B~\cite{chen2024internvl}} & \textbf{80.15} & \textbf{66.39} & 72.04 & 62.18 & \textbf{80.01} & 78.69 & 82.14 & 91.31 & 79.91 & 70.65 & 85.62 & 86.74 & 64.18 & 76.92 \\          
         \scalebox{0.8}[0.8]{Qwen2-VL-72B~\cite{wang2024qwen2}} & 79.21 & 65.49 & 69.77 & 55.87 & 60.33 & 84.39 & 88.91 & 95.8 & 81.59 & 75.08 & 85.37 & 88.61 & 82.9 & 77.95 \\         
         \scalebox{0.8}[0.8]{Gemini-1.5-Pro~\cite{team2023gemini}} & 79.76 & 62.97 & \textbf{78.83} & \textbf{74.83} & 64.05 & 85.57 & \textbf{94.77} & 95.52 & \textbf{90.51} & \textbf{84.07} & 86.44 & \textbf{92.58} & \textbf{92.31} & \textbf{83.25} \\
         \bottomrule
         
    \end{tabulary}

    \label{tab:result_conventional_ocr}
    \vspace{-3mm}

\end{table*}
In this section, we evaluate nine representative LMMs either with open-source models or commercial APIs. The commercial APIs with specific versions are GPT-4o-2024-08-06, Gemini-1.5-Pro-002, Claude-3.5-Sonnet-20241022. The open-source LMMs include KOSMOS2.5, TextMonkey, Florence, GOT, InternVL2-76B, and Qwen2-VL-72B. 

The overall evaluation results are shown in Fig.~\ref{fig:overall_performance}. It can be observed that Gemini-1.5-Pro achieves the highest ranking with an average score of 73.0, securing first place in three tracks. This demonstrates its outstanding and robust OCR capabilities. Qwen2-VL-72B ranks second overall, with an average score of 68.7, and notably ranks first in the KIE track. The results indicate that generalist models tend to perform better than specialized ones.


\subsection{Multi-Scene OCR}
We evaluate the generic LMMs and OCR-specialized LMMs on the track of multi-scene OCR. As presented in Tab.~\ref{tab:result_conventional_ocr}, generic LMMs models significantly outperform specialized models, primarily due to their huge parameter and training data. Among all the models, Gemini-1.5-Pro performs the best with a score of 83.25, followed by Qwen2-VL-72B with a score of 77.95. In the category of specialized models, GOT is the top performer with a score of 61.00. Across different scenes, it is evident that the performance exhibits considerable variation. For most models, the average performance in natural scenes is approximately 15\% lower compared to their results in document scenes.

\begin{table*}[]
\small
\centering
\caption{Performance of LMMs on subsets of Multilingual OCR track.}
\begin{tabular}{cccccccccccc}
\toprule
Method & \scalebox{0.9}[0.9]{Korean} & \scalebox{0.9}[0.9]{Japanese} & \scalebox{0.9}[0.9]{Vietnamese} & \scalebox{0.9}[0.9]{French} & \scalebox{0.9}[0.9]{German} & \scalebox{0.8}[0.9]{Italian} & \scalebox{0.9}[0.9]{Spanish} & \scalebox{0.9}[0.9]{Portuguese} & \scalebox{0.9}[0.9]{Russian} & \scalebox{0.9}[0.9]{Arabic} & \scalebox{0.9}[0.9]{Total} \\          \midrule
GOT & 27.53	& 43.66	& 7.92 & 33.74 & 28.73 & 28.53 & 31.63 & 25.20 & 3.85 & 18.72 & 24.95 \\
KOSMOS2.5 & 26.95 & 26.94 & 27.24 & 48.93 & 47.04 & 44.59 & 52.92 & 54.56 & 11.13 & 22.04 & 36.23 \\ 
Florence & 27.93 & 34.76 & 21.96 & 70.21 & 61.30 & 61.62 & 70.44 & 67.78 & 26.74 & 54.24 & 49.70 \\
\midrule
InternVL2-76B & 31.58 & 44.52 & 20.07 & 66.68 & 58.5 & 65.78 & 71.51 & 70.41 & 13.18 & 23.52 & 46.57 \\
Claude-3.5-Sonnet & 58.72 & 56.9 & 59.58 & 76.76 & 70.66 & 68.38 & 73.42 & 76.07 & 51.01 & 65.29 & 65.68 \\
Qwen2-VL-72B & 69.13 & 62.28 & 64.37 & 80.34 & 73.34 & 69.6 & 77.58 & 80.31 & 65.94 & 68.48 & 71.14 \\
GPT-4o & 74.2 & 66.96 & 70.11 & 81.17 & 73.6 & 69.01 & 78.95 & 80.9 & 67.22 & 72.31 & 73.44 \\
Gemini-1.5-Pro & \textbf{80.01} & \textbf{73.52} & \textbf{78.49} & \textbf{83.33} & \textbf{78.11} & \textbf{75.77} & \textbf{81.28} & \textbf{83.46} & \textbf{69.99} & \textbf{85.7} & \textbf{78.97} \\ 
\bottomrule
\end{tabular}

\label{tab:result_multi_lan}
\end{table*}

\begin{table*}[]
\centering
\caption{Performance of LMMs on Document Parsing track. * denotes that we convert the outputs of the model to HTML format.}
\begin{tabular}{cccccccccccc} \toprule
  \multirow{3}{*}{Method}           & \multicolumn{4}{c}{Document}                         & \multicolumn{4}{c}{Table}                            & \multicolumn{2}{c}{Formula}  &\multirow{3}{*}{Total} \\ \cmidrule(lr){2-5} \cmidrule(lr){6-9} \cmidrule(lr){10-11}
             & \multicolumn{2}{c}{scan} & \multicolumn{2}{c}{photo} & \multicolumn{2}{c}{scan} & \multicolumn{2}{c}{photo} & \multicolumn{2}{c}{handwritten}  & \\ \cmidrule(lr){2-3} \cmidrule(lr){4-5} \cmidrule(lr){6-7} \cmidrule(lr){8-9} \cmidrule(lr){10-11}
             & en         & zh        & en          & zh        & en         & zh        & en         & zh         & math       & molecular  &    \\
             \midrule
InternVL2-76B   &24.61&16.42&49.42&35.49&64.11&40.63&48.14&30.26&28.33&15.92 &35.33\\
GOT  &65.39&\textbf{69.91}&64.07&61.84&47.01*&20.67*&16.90*&11.57*&14.31&20.16&39.18\\ 
Claude-3.5-Sonnet &46.85&30.77&60.63&27.39&76.50&49.95&63.67&41.47&\textbf{43.98}&36.65&47.79\\
GPT-4o &64.19&52.33&65.72&41.32&74.72&57.59&65.89&42.73&41.64&26.83&53.30\\
Qwen2-VL-72B  &\textbf{66.51}&62.36&\textbf{69.84}&60.74&65.96&49.76&59.34&41.15&41.58&20.54&53.78\\ 
Gemini-1.5-Pro &64.69&66.01&69.38&\textbf{68.63}&\textbf{80.99}&\textbf{64.55}&\textbf{65.92}&\textbf{60.26}&27.17&\textbf{56.14}&\textbf{62.37}\\
\bottomrule

\end{tabular}

\label{tab:exp_doc_parsing}
\vspace{-2mm}
\end{table*}

\subsection{Multilingual OCR}
As shown in Tab.~\ref{tab:result_multi_lan}, the performance gap among models in the multilingual track is more notable. Among the generalist models, Gemini-1.5-Pro ranks first with a score of 78.97. In the specialized group, Florence is the best and achieves a score of 49.70. Additionally, the table illustrates a significant variation across different languages for every model. The accuracy of languages in Asia, such as Japanese, Korean, Vietnamese, and Arabic, is generally lower than languages in the Latin family. Among the Asian languages, Japanese exhibits the lowest performance for most generalist models, which we attribute to the higher prevalence of vertical text in our collected images. In the Latin family, German and Italian often exhibit lower performance, possibly due to the presence of more special characters. Overall, the multilingual capabilities of most models are inferior to their performance in Chinese and English with the results presented in Tab.~\ref{tab:result_conventional_ocr}. Therefore, improving multilingual capabilities is a future direction for most models. 
\subsection{Document Parsing}
In the document parsing track, most specialized models except GOT do not support LaTeX or HTML formats. Therefore, we primarily evaluate the generalist models, and the results are presented in Tab.~\ref{tab:exp_doc_parsing}. Gemini-1.5-Pro, Qwen2-VL-72B, and GPT-4o demonstrate superior performance across various document parsing subsets, with Gemini-1.5-Pro achieving the best performance, scoring 62.37.  

However, a detailed analysis of individual sub-sets reveals that the top-performing large models still attain average scores below 70. For instance, Gemini-1.5-Pro achieves an average score of 67.17 in document content structuring and 67.93 in table recognition, indicating considerable room for improvement. This also demonstrates that our document parsing benchmark is comprehensive and challenging. More challenging document image examples can be found in Appendix.~\ref{sec:vis_data}.

Furthermore, we observe that most models perform better on English than Chinese documents. For example, Gemini-1.5-Pro scores 5\% higher on English tables than on Chinese tables. The gap indicates that there is an imbalance in the preparation of training data across different languages.

It is noteworthy that all models perform particularly poorly on handwritten formulas, with scores rarely exceeding 50. The only exception is Gemini-1.5 Pro on molecular formula is an exception. This indicates that there is significant room for improvement in the recognition of handwritten formulas.

\begin{table*}[]
\centering
\vspace{-2mm}
\caption{Results of LMMs on subsets of KIE track.}
\begin{tabular}{cccccccc}
\toprule
\multirow{2}{*}{Method} & \multicolumn{4}{c}{Constrained Category}     & \multicolumn{2}{c}{Open Category} & \multirow{2}{*}{Total} \\  \cmidrule(lr){2-5} \cmidrule(lr){6-7}
    & SROIE & CORD & EPHOIE & POIE & SOLD-SIBR       & SOLD-HUST-CELL     &                        \\ \toprule
    InternVL2-76B & 57.06 & 66.12 & 59.03 & \textbf{71.09} & 50.08 & 66.22 & 61.60  \\ 
    GPT-4o  & 72.55 & \textbf{81.26} & 44.12 & 59.01 & 63.60 & 66.94 & 63.45 \\
    Claude-3.5-Sonnet & 69.83 & 71.51 & 42.82 & 59.25     & 66.20 & 71.07 & 64.58 \\
    Gemini-1.5-Pro & 75.00 & 78.73 & 61.27 & 61.49     & 59.47 & 67.72 & 67.28 \\

    Qwen2-VL-72B & \textbf{75.50} & 69.02     & \textbf{69.98} & 62.47 & \textbf{72.43} & \textbf{81.17} & \textbf{71.76}                 \\

 \bottomrule
\end{tabular}

\label{tab:result_kie}
\vspace{-1mm}
\end{table*}

\subsection{Key Information Extraction}
The KIE task requires models to output structured JSON, which reduces the number of comparable models to only five generalist models.  As presented in ~\cref{tab:result_kie}, the large models exhibit robust performance across both constrained-category and open-category KIE tasks. Qwen2-VL-72B and Gemini-1.5-Pro rank first and second, with a score of 71.76 and 67.28 respectively. Specifically, Gemini-1.5-Pro outperforms Qwen2-VL-72B by 19.71\% on the English test set CORD. Conversely, Qwen2-VL-72B shows a comparative advantage on the open-category sub-sets, particularly on SOLD-SIBR and SOLD-HUST-CELL, significantly surpassing other models in performance. The GPT-4o and Claude-3.5-Sonnet exhibit suboptimal results on the EPHOIE dataset. Meanwhile, InternVL2, an excellent open-source model, performs best on the POIE dataset but less effectively on the initial releases of the SOLD-SIBR and SOLD-HUST-CELL datasets.  The highest-performing model in this evaluation achieves an F1-score of only 71.76\% indicating that its performance still falls short of practical application. This can be attributed to the challenge of the dataset, which includes factors such as real-world noise, complex structures, and the requirement for JSON-formatted output.
\vspace{-2mm}
\subsection{Analysis}
In addition to the four main experiments, we conduct additional analyses to uncover the limitations of the current LMMs. It can be observed that the fine-grained image perception capabilities of current large models are generally weaker compared to their recognition abilities. For example, most models lack grounding capabilities, and their performance tends to decrease when images are rotated.

\subsubsection{Visual Text Grounding}



\begin{table*}[h]
    \centering
    \caption{Comparison of text grounding performance on English subsets from Multi-Scene OCR.}    
    \small
    \begin{tabular}{cccccccccc}
        \toprule
         \multirow{2}{*}{Method} & \multicolumn{4}{c}{Natural Scene OCR} & \multicolumn{3}{c}{Document OCR} & \multicolumn{1}{c}{UGC Image OCR} & \multirow{2}{*}{Total} \\
         \cmidrule(lr){2-5} \cmidrule(lr){6-8} \cmidrule(lr){9-9}
           & TotalText & IC15 & InverseText & HierAgent & FUNSD & CORD & IAM & LAION-OCR &  \\
         \midrule
         KOSMOS2.5 & 17.6 & 28.09 & 8.55 & 20.64 & 13.59 & 31.59 & 25.12 & 25.12 & 21.29  \\
         Florence & 27.02 & 31.38 & 22.49 & 23.35 & 8.85 & 43.52 & 0.2 & 37.09 & 24.24 \\
         TextMonkey & 58.87 & 35.17 & 38.15 & 31.37 & 9.42 & 69.94 & \textbf{58.39} & 73.71 & 46.95 \\
         Gemini-1.5-Pro & \textbf{67.25} & \textbf{52.00} & \textbf{66.77} & \textbf{47.74} & \textbf{43.82} & \textbf{76.48} & 56.05 & \textbf{77.71} & \textbf{60.98} \\         
         \bottomrule
         
    \end{tabular}

    \vspace{-3mm}
    \label{tab:result_text_grounding}
\end{table*}
We explore the text grounding capabilities of LMMs, which are widely applicable in various scenarios such as image translation, text editing, and text-centric object detection.
We evaluate the grounding capabilities using English subsets from the multi-scene OCR track, and the results are presented in Tab.~\ref{tab:result_text_grounding}. The evaluation metrics are introduced in Sec.~\ref{sec:details_eval}. We observe that only four models support text grounding, and their performance in text grounding decreases significantly compared to their text recognition results. Gemini-1.5-Pro achieves the highest score among all the models with a score of 60.98\%, while the other models do not exceed 50\%. These results demonstrate that fine-grained text grounding remains a major challenge for current LMMs.

\begin{table}[]
    \centering
    \caption{Comparsion of results on the normal and multi-oriented benchmark of Multi-Scene OCR task with English and Chinese. Note that M.O. stands for multi-oriented benchmark.}    
    \begin{tabular}{cccc} \toprule
         \multirow{2}{*}{Method} &  \multicolumn{2}{c}{Total} & \multirow{2}{*}{Diff} \\
         \cmidrule(lr){2-3} 
         & Normal & M.O. & \\
         \midrule
         GOT & 61.00 & 26.14 & -34.86 \\  
         KOSMOS2.5& 47.55 & 28.84 & -18.71 \\
         \midrule
         InternVL2-76B & 76.92 & 38.00 & -38.92 \\
         Claude-3.5-Sonnet & 72.87 & 42.42 & -30.45 \\         
         Qwen2-VL-72B & 77.95 & 54.62 & -23.33 \\
         GPT-4o & 76.40 & 57.02 & -19.38 \\         
         Gemini-1.5-Pro & 83.25 & 79.45 & \textbf{-3.80} \\
         \bottomrule
         
    \end{tabular}
    \vspace{-4mm}
    \label{tab:result_multi_oriented_ocr}
\end{table}

\subsubsection{Multi-Orientation}
Multi-orientation text is a challenge in traditional OCR. Therefore, we also evaluate the robustness of large models in handling multi-orientation text, and the experimental results are presented in Tab.~\ref{tab:result_multi_oriented_ocr}.
We rotate the images of the multi-scene OCR benchmark by 0, 90, 180, or 270 degrees to build a multi-orientation benchmark. It can be observed that most methods perform much worse on multi-orientation images than on original images.
Claude-3.5-Sonnet, GOT, Qwen2-VL-72B, and InternVL2 show a performance decrease of more than 23\%. GPT-4o and KOSMOS2.5 result in a reduction of around 19\%, while Gemini-1.5-Pro exhibits only a slight decline of 3.80\%. These results emphasize the need for the inclusion of more rotated training data to mitigate this challenge.


\subsubsection{Hallucination of Repetition Output}

Based on our observation, most large models tend to produce hallucinations in the form of repeated text during prediction. So we devise a repetition ratio $R_{rep}$ by measuring the number of images producing repetitive output out of the overall number of images. Detailed introduction of $R_{rep}$ can be referred to Appendix.~\ref{sec:details_eval}. The repetition performance is presented in Tab.~\ref{tab:result_repetition}. We observe that generalist LMMs produce less repetition results than specialized methods. Among all the models, Claude-3.5-Sonnet has the lowest repetition rate, with only one in a thousand images exhibiting repeated outputs. TextMonkey, on the other hand, has the most severe hallucination of repetition, with a rate of 33.93\%. 
\section{Conclusions and Future Works}

\begin{table}[]
    \centering
    \caption{Hallucination of repetition of LMMs on data from Multi-Scene OCR track. $R_{rep}$ denotes the rate of Repetition.}    
    \begin{tabular}{cc} \toprule
         Method & $R_{rep}$($\downarrow$) \\
         \midrule
         InternVL2-76B & 5.94\%  \\      
         Qwen2-VL-72B & 0.65\% \\
         GPT-4o &  0.28\% \\         
         Gemini-1.5-Pro & 0.14\%  \\      
         Claude-3.5-Sonnet & 0.09\%  \\
         \midrule         
         TextMonkey & 33.93\%  \\
         KOSMOS2.5 & 10.64\%  \\
         GOT & 2.72\%  \\
         Florence & 0.68\% \\         
         \bottomrule
         
    \end{tabular}

    \vspace{-2mm}
    \label{tab:result_repetition}
\end{table}

In this paper, we introduce CC-OCR, a comprehensive and challenging benchmark designed to evaluate the visual literacy and document processing capabilities of state-of-the-art LMMs. By covering a diverse range of scenarios, tasks, and challenges, CC-OCR addresses critical gaps in existing benchmarks, offering a holistic evaluation framework for multilingual text reading, multi-scene OCR, document parsing, and key information extraction. Through rigorous experiments on nine prominent models, including five generalist and four specialized LMMs, we underscore the importance of handling fine-grained text grounding, multi-oriented and multilingual text, handwritten formulas, and hallucinations of repetition errors to advance the literate intelligence of LMMs. We hope that CC-OCR will serve as a valuable and challenging benchmark for the research community, driving progress in the development of more capable and reliable LMMs for diverse real-world OCR tasks.
\clearpage

{
    \small
    \bibliographystyle{ieeenat_fullname}
    \bibliography{main}
}

\clearpage
\appendix

\section{Appendix}
\subsection{Detailed Composition of \ccocr}
\label{sec:details_ccocr}
The following Tab.~\ref{tab:composition_cc_ocr} provides detailed statistics of the \ccocr benchmark. The benchmark includes four primary tracks and diverse subsets to cover various challenges comprehensively.

\textbf{Multi-Scene OCR} track consists of a total of 2750 images and we split them into three subsets: Natural Scene, Document, and UGC Image OCR with 1850, 350, and 550 samples for each subset. All subsets contain both Chinese and English scripts. For Natural Scene OCR, the images are carefully selected from both academic datasets (TotalText, IC15, InverseText, and HierText) and our own collected data. For Document OCR, the images are mainly selected from our own collected documents of various types including printed documents, real-shot documents, handwritten documents, and more. For Web Image OCR, we select images from screenshots of text-rich websites and user-generated content (UGC) images.

\textbf{Multilingual OCR} track consists of a total of 1500 images with 150 samples for each language. This track includes two East Asian languages (Japanese and Korean), six Latin alphabet-based languages (French, German, Italian, Spanish, Portuguese, and Vietnamese), one Cyrillic alphabet-based language (Russian), and one Arabic alphabet-based language (Arabic). Similarly to the multi-scene OCR track, images are selected from various scenarios including natural scenes, documents, and handwritten images.

\textbf{Document Parsing} track consists of a total of 800 images with 150 samples for printed documents, 150 samples for real-shot samples, 300 samples for table parsing, 100 for mathematical expression recognition, and 100 for molecular formula recognition. The document images contain a wide variety of elements, including figures, charts, tables, and mathematical expressions.

\textbf{Key Information Extraction} track consists of 2008 images, divided into two groups: constrained-category subset with 1008 images and open-category subset with 1000 images. The constrained-category subset includes 347 images for SROIE, 100 for CORD, 311 for EPHOIE, and 250 for POIE. The EPHOIE data is in Chinese, and the others are in English. The 1000-image open-category set, namely COLD-KIE data, contains 400 images from SIBR and 1426 images from HUST-CELL that remain after filtering the pure table images, and it contains both Chinese and English scripts. Among all the data, the ratio of English to Chinese is 1:1.48.

\begin{table*}[ht]
    \small
    \centering
        \begin{tabular}{c|c|c|c|c|c|c}
          \hline
          \multirow{2}{*}{Dataset} & \multirow{2}{*}{Subset} & \multirow{2}{*}{Track} & \multirow{2}{*}{Feature} & \multirow{2}{*}{\makecell{Evalutaion\\Granularity}} & \multirow{2}{*}{Source} & \multirow{2}{*}{Images} \\
          & & & & & \\
          \hline
          TotalText & \multirow{5}{*}{Natural Scene} & \multirow{13}{*}{Multi-Scene OCR} & curved & word & open-source & 300 \\
          \cline{1-1}
          \cline{4-7}
          IC15 & & & blurred & word & open-source & 500 \\
          \cline{1-1}
          \cline{4-7}
          InverseText & & & inverse & word & open-source & 500 \\
          \cline{1-1}
          \cline{4-7}
          HierAgent & & & dense & word & open-source &100 \\
          \cline{1-1}
          \cline{4-7}
          Scene\_zh & & & scene & char & collected &450 \\
          \cline{1-2}
          \cline{4-7}
          FUNSD & \multirow{5}{*}{Document} & & doc & word & open-source &50 \\
          \cline{1-1}
          \cline{4-7}
          CORD & & & doc & word & open-source &100 \\
          \cline{1-1}
          \cline{4-7}
          IAM & & & handwritten & word & open-source &50 \\
          \cline{1-1}
          \cline{4-7}
          Doc\_zh & & & doc & char & collected &100 \\
          \cline{1-1}
          \cline{4-7}
          Doc\_hw\_zh & & & handwritten & char & collected &50 \\
          \cline{1-2}
          \cline{4-7}
          UGC\_ver\_zh & \multirow{3}{*}{UGC Image} & & vertical & char & collected &100 \\
          \cline{1-1}
          \cline{4-7}
          DAST\_zh & & & dense & char & open-source &50 \\
          \cline{1-1}
          \cline{4-7}
          LAION-OCR & & & artistic & word & open-source &400 \\
          \cline{1-3}
          \cline{4-7}
          Arabic & Arabic & \multirow{10}{*}{Multilingual OCR} & \multirow{10}{*}{\makecell{scene,doc, \\handwritten}}& word & \multirow{10}{*}{\makecell{open-source\\(re-annotated)\\ \& collected}} &150 \\
          \cline{1-2}
          \cline{5-5}
          \cline{7-7}
          Korean & \multirow{2}{*}{East Asia} & & & char & &150 \\
          \cline{1-1}
          \cline{5-5}
          \cline{7-7}
          Japanese & & & & char &  &150 \\
          \cline{1-2}
          \cline{5-5}
          \cline{7-7}
          French & \multirow{6}{*}{Latin}& & & word & &150 \\
          \cline{1-1}
          \cline{5-5}
          \cline{7-7}
          German & & & & word & &150 \\
          \cline{1-1}
          \cline{5-5}
          \cline{7-7}
          Italian & & & & word & &150 \\
          \cline{1-1}
          \cline{5-5}
          \cline{7-7}
          Spanish & & & & word & &150 \\
          \cline{1-1}
          \cline{5-5}
          \cline{7-7}
          Portuguese & & & & word & &150 \\
          \cline{1-1}
          \cline{5-5}
          \cline{7-7}
          Vietnamese & & & & word & &150 \\
          \cline{1-2}
          \cline{5-5}
          \cline{7-7}
          Russian & Cyrillic & & & word & &150 \\
          \cline{1-7}
          Doc\_scan\_en & \multirow{4}{*}{\makecell{Full-page\\Document}} & \multirow{10}{*}{Doc Parsing} &\multirow{4}{*}{intricate, diverse} & \multirow{10}{*}{\makecell{full-page\\edit-distance}} & \multirow{8}{*}{collected} &75 \\
          \cline{1-1}
          \cline{7-7}
          Doc\_scan\_zh & & & &  & &75 \\
          \cline{1-1}
          \cline{7-7}
          Doc\_photo\_en & & &  & & &75 \\
          \cline{1-1}
          \cline{7-7}
          Doc\_photo\_zh &  & &  & & &75 \\
          \cline{1-2}
          \cline{4-4}
          \cline{7-7}
          Table\_scan\_en & \multirow{4}{*}{Table} & &\multirow{4}{*}{complex, long} &  & &75 \\
          \cline{1-1}
          \cline{7-7}
          Table\_scan\_zh &  &  & & & &75 \\
          \cline{1-1}
          \cline{7-7}
          Table\_photo\_en &  &  & & & &75 \\
          \cline{1-1}
          \cline{7-7}
          Table\_photo\_zh & & & & & &75 \\
          \cline{1-2}
          \cline{4-4}
          \cline{6-7}
          Math & \multirow{2}{*}{Formula} & &\multirow{2}{*}{handwritten}  & & collected &100 \\
          \cline{1-1}
          \cline{7-7}
          Molecular &  & & & &  open-source &100 \\
          \cline{1-2}
          \cline{7-7}
          \hline
          SROIE & \multirow{4}{*}{\makecell{Constrained\\Category}} & \multirow{6}{*}{Key Information Extraction} & \multirow{6}{*}{\makecell{scene, \\handwritten \\hierarchical, \\complex tables}} & \multirow{6}{*}{entity} & \multirow{4}{*}{open-source} & 347 \\
          \cline{1-1}
          \cline{7-7}
          CORD & & & & &  &100 \\
          \cline{1-1}
          \cline{7-7}
          EPHOIE & & & &  & &311 \\
          \cline{1-1}
          \cline{7-7}
          POIE &  & & &  & &250 \\
          \cline{1-2}
          \cline{6-7}
          SOLD-SIBR & \multirow{2}{*}{Open Category} & & &  & \multirow{2}{*}{\makecell{open-source\\(re-annotated)}} & 400 \\
          \cline{1-1}
          \cline{7-7}
          SOLD-HUST-CELL &  & &  & & &600 \\
          \cline{1-1}
          \cline{7-7}
          \hline
          Total & - & - & - & - & - & 7058 \\
          \hline
          \end{tabular}
      \caption{Details of the \ccocr Benchmark.}
      \label{tab:composition_cc_ocr}
  \end{table*}

\subsection{Detailed Annotation of KIE}
\label{sec:details_anno_kie}
SIBR and HUST-CELL use various types of links or identifier associations to represent complex structures.  However, the links among entities are local, short, and disordered, without taking the global semantic information into account, which may lead to key conflicts when building end-to-end representation. In addition, the key-value pairs in the table regions lack hierarchical information and cannot form rows.

\begin{figure}[h]
    \includegraphics[width=0.8\columnwidth]{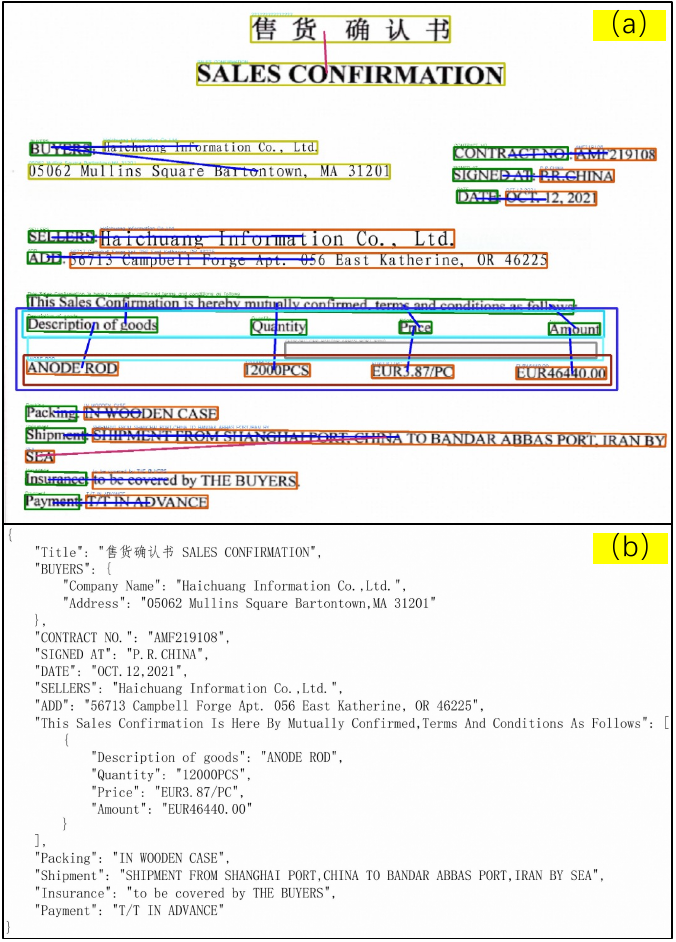}
    \centering
    \caption{The dataset annotation demonstration of key information extraction. The sub-image (a)  is the visualization of data annotation, where the green, orange, and yellow boxes represent key, value, and independent entity, respectively. The blue, cyan, and red boxes represent the table region, table header, and table row. The sub-image (b) shows the end-to-end JSON representation parsed according to the bottom-up annotations. }
    \label{fig:vie_challenges}
\end{figure}

To handle this problem, we relabel the images in the SIBR and HUST-CELL with a bottom-up annotation approach, resulting in an end-to-end representation. Specifically, leveraging SIBR entity annotation and utilizing both entities and external links to establish relationships, we have addressed the following issues: a) Multilevel Link Conflicts: We transformed links into directed, ordered long links to resolve hierarchical and complex text reading order challenges. b) Annotation of Keyless Values: We annotate the corresponding categories for such values as keys, including independent entities, untitled tables, etc. c) Table Structure Annotation: We annotate the table header, row numbers, and table footer to facilitate accurate table parsing.  The annotation process is shown in Fig.~\ref{fig:vie_challenges} and is described as follows: First, we filter out pure table data and low-quality data. The images are then annotated with rule-based methods for multiple rounds of quality rectification, such as detecting key conflicts. Finally, the post-processing adopted includes: a) error correction, we use LLMs to correct errors in annotations; b) Key Normalization, the spaces between Chinese characters are removed, and symbols are changed to half-width; c) parsing into end-to-end JSON format. All tasks involving LLMs undergo secondary manual checks to ensure data quality.

\subsection{Detailed introduction of Evaluation Method}
\label{sec:details_eval}
\subsubsection{OCR}
To comprehensively evaluate the OCR capabilities of Large Vision-Language Models (LVLMs) in various scenarios, and considering that most large models have different orders to predict text sequences, we propose using full-text multi-set matching to assess their OCR capabilities. Specifically, we first split ground-truth $\mathrm{T}$ and predicted text sequences $\mathrm{T}^{'}$ into two sets of basic units $\mathrm{U}=\{(u_0, c_0), ..., (u_N, c_N)\}$ and $\mathrm{U}^{'}=\{(u_0, c^{'}_i),...,(u_N, c^{'}_N)\}$. $u_i$, $c_i$, and $c^{'}_i$ represent a basic unit, its count in the gt sequence, and its count in the predicted sequence. $N$ indicates the size of the union of gt basic units and predicted basic units. Then, we generate both recall and precision metrics with the following protocol and assess their performances with F1-score: 

\begin{equation}
    Recall = \frac{\sum_1^{N}min(c_i, c^{'}_i)}{\sum_1^{N}c_i}
\end{equation}

\begin{equation}
    Precision = \frac{\sum_1^{N}min(c_i, c^{'}_i)}{\sum_1^{N}c^{'}_i}
\end{equation}

Two basic units are considered `matched' if and only if they are identical.

For different languages, we have designed targeted splitting protocols to define a basic unit. For instance, for Chinese, Japanese, and Korean (CJK) languages, we define characters as the basic units, whereas for Latin, Cyrillic, and Arabic scripts, we adopt words as the basic units for evaluation.

For evaluation of text grounding performance, following the previous works~\cite{liu2024textmonkey}~\cite{liu2023spts}, we evaluate the text grounding ability with point-based evaluation metric of text spotting. That is, we first match all words with ground truth with a point-based distance metric, i.e., the predicted point (or central point of predicted boxes ) that has the nearest distance to the central point of the GT box would be selected, and the recognition results will be measured by the same full-matching rules used in existing benchmarks. Only one predicted point with the highest confidence will be matched to the ground truth. And the predicted point or box is defined as \textit{True Positives} (i.e. TP), when it is matched to the ground truth and the predicted text, is identical to the ground truth, and the other predictions are then marked as \textit{False Positives} (i.e. FP), the other ground truth is marked as \textit{False Negatives} (i.e. FN). Thus, recall and precision of text grounding are measured as follows.

\begin{equation}
Recall = \frac{\text{TP}}{\text{TP} + \text{FN}}
\label{eq:grounding_r}
\end{equation}

\begin{equation}
Precision = \frac{\text{TP}}{\text{TP} + \text{FP}}
\label{eq:grounding_p}
\end{equation}
The F1-score is calculated by combining recall and precision.

\subsubsection{Document Parsing}
The datasets for the various document parsing subtasks exhibit significant differences in element representations and annotation formats. To comprehensively and accurately evaluate the document parsing capabilities of different models, we have developed distinct evaluation methods tailored to each sub-dataset.

For document content structuring, which primarily involves plain text, formulas, and tables, we assess the similarity between model predictions and the ground truth using the normalized edit distance (NED)~\cite{lcvenshtcin1966binary}. Specifically, as shown in~\cref{eq:ned}, \( N \) represents the number of images in the dataset. The functions \( \text{EditDist}(\cdot) \) and \( \text{len}(\cdot) \) denote the edit distance and the length of a string, respectively. \( \text{P}_i \) and \( \text{G}_i \) represent the predicted and ground truth results for the \( i \)-th sample. The normalized edit distance effectively reflects the model's ability to structure document content.

\begin{equation}
    \text{NED} = \frac{1}{N} \sum_{i=1}^{N} \left( 1 - \frac{\text{EditDist}(\text{P}_i, \text{G}_i)}{\max(\text{len}(\text{P}_i), \text{len}(\text{G}_i))} \right)
    \label{eq:ned}
\end{equation}

For the table recognition task, the dataset annotations are provided in HTML format. We utilize Tree Edit Distance-based Similarity (TEDS)~\cite{zhong2020image} to measure the differences between predictions and ground truth. TEDS accounts for both the structural similarity of tables and the accuracy of the predicted cell contents, effectively evaluating the model's table recognition performance. As illustrated in \cref{eq:teds}, let \( T_{pred} \) and \( T_{gt} \) denote pred and gt trees encoded in HTML respectively, with \( |T_{pred}| \) and \( |T_{gt}| \) representing the number of nodes in trees \( T_{pred} \) and \( T_{gt} \), respectively. The edit distance, expressed as \( \text{EditDist}(T_{pred}, T_{gt}) \), signifies the minimum number of node edit operations required to convert \( T_{pred} \) into \( T_{gt} \). Similar to document content structuring, we use the normalized TEDS score at the dataset level as the evaluation metric.

\begin{equation}
    \text{TEDS}(T_{pred}, T_{gt}) = 1 - \frac{\text{EditDist}(T_{pred}, T_{gt})}{\max(|T_{pred}|, |T_{gt}|)}
    \label{eq:teds}
\end{equation}

\begin{equation}
    \text{NTEDS} = \frac{1}{N} \sum_{i=1}^{N} \left( TEDS_i \right)
    \label{eq:nteds}
\end{equation}

For handwritten formulas and chemical molecular formulas, we apply the same evaluation method used for document content structuring, utilizing the normalized edit distance as the performance metric. This approach ensures consistency in assessing document parsing capabilities across different subtasks.

\subsubsection{Key Information Extraction}
Following the previous works~\cite{kim2022ocr,hwang2019post}, we evaluate the performance of the KIE task with a field-level F1 score. As defined as \cref{eq:f1}, the \textit{True Positives}, \textit{False Positives}, \textit{False Negatives} are abbreviated as TP, FP, FN. The F1 checks whether the extracted field information is in the ground truth, where a key-value pair is treated as a field, and even if a single character is missed, the score assumes the field extraction is failed. 
In addition, during the evaluation, we removed some spaces that do not affect semantics and word tokenization. For example, if the model predicted value is ``3.5 kg", the true value is ``3.5kg" and it will be considered correct. The normalization code will be released.

\begin{equation}
F1 = \frac{2 \times \text{TP}}{2 \times \text{TP} + \text{FP} + \text{FN}}
\label{eq:f1}
\end{equation}

\subsubsection{Repetition}
We devise a repetition ratio $Rrep$ to represent the repetition performance, by measuring the number of images producing repetitive output out of the overall number of images. In detail, an image is tagged as image producing repetitive output, when the ratio of the length of the continuous repetitive string of prediction to the full length of prediction is larger than 0.25. The continuous repetitive string is defined when the frequency of every continuous unit (divided by space or newline character) exceeds a threshold $T_{rep}$, and $T_{rep}$ is set to 5 experimentally.

\subsection{Visualization of More Data}
\label{sec:vis_data}
In this section, we present more examples of the proposed \ccocr.

\subsubsection{hallucination}

\begin{figure}[h]
    \includegraphics[width=\columnwidth]{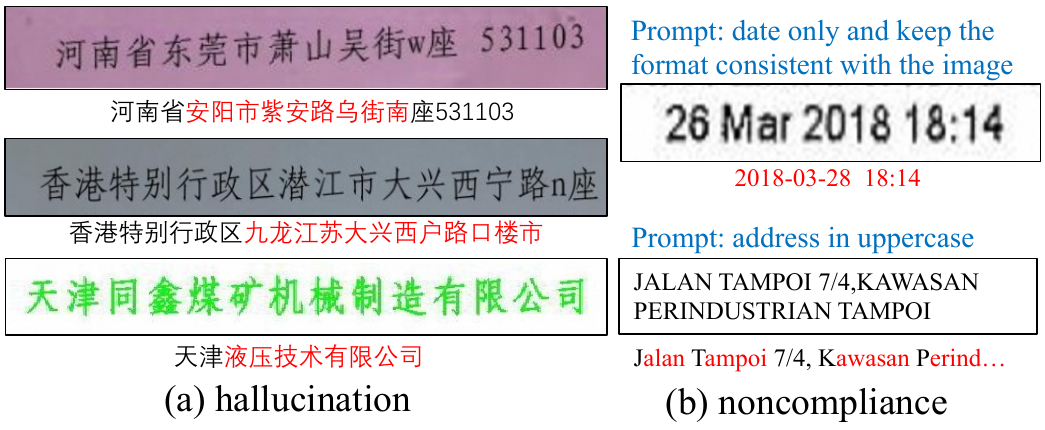}
    \centering
    \caption{Visualization of Hallucination and Noncompliance in Vision-Language Models for KIE Tasks. The red characters denote errors, whereas the blue characters indicate the prompt.}
    \label{fig:kie_finding}
\end{figure}

In our analysis of the failure cases encountered by GPT-4o in the KIE task, we have made several observations: 1) Hallucination: For lengthy textual information, such as addresses and company names, it appears that large language models struggle to maintain strict consistency with the content presented in the images. This inconsistency often results in content tampering, as illustrated in ~\cref{fig:kie_finding}-(a). Specifically, the model tends to modify the content in long list areas, wherein non-existent values might appear. These inserted values generally align with the overall semantic context. 2) Noncompliance: the large language models might fail to adhere to instructions fully. For instance, when evaluating the SROIE dataset, we specify that addresses must be in uppercase, and dates should be presented without specific times. However, empirical evaluations reveal that not all samples conform stringently to these prompts, as demonstrated in ~\cref{fig:kie_finding}-(b). A recurring issue with GPT-4o is its tendency to standardize time formats, such as transforming "26 May 2018" into "2018-05-26".

\subsubsection{Examples of Multi-scene OCR}

\begin{figure*}[t]\centering

        

        
        
\begin{minipage}{0.8\linewidth}\vspace{0mm}    
    \centering
    \scriptsize
    \begin{tcolorbox}[boxrule=0.2mm]
        \centering
        \hspace{-5mm}
        \begin{tabular}{p{0.99\columnwidth}}
        \hspace{1mm}
        \begin{minipage}{0.99\columnwidth}
        
        \includegraphics[width=.6\linewidth]{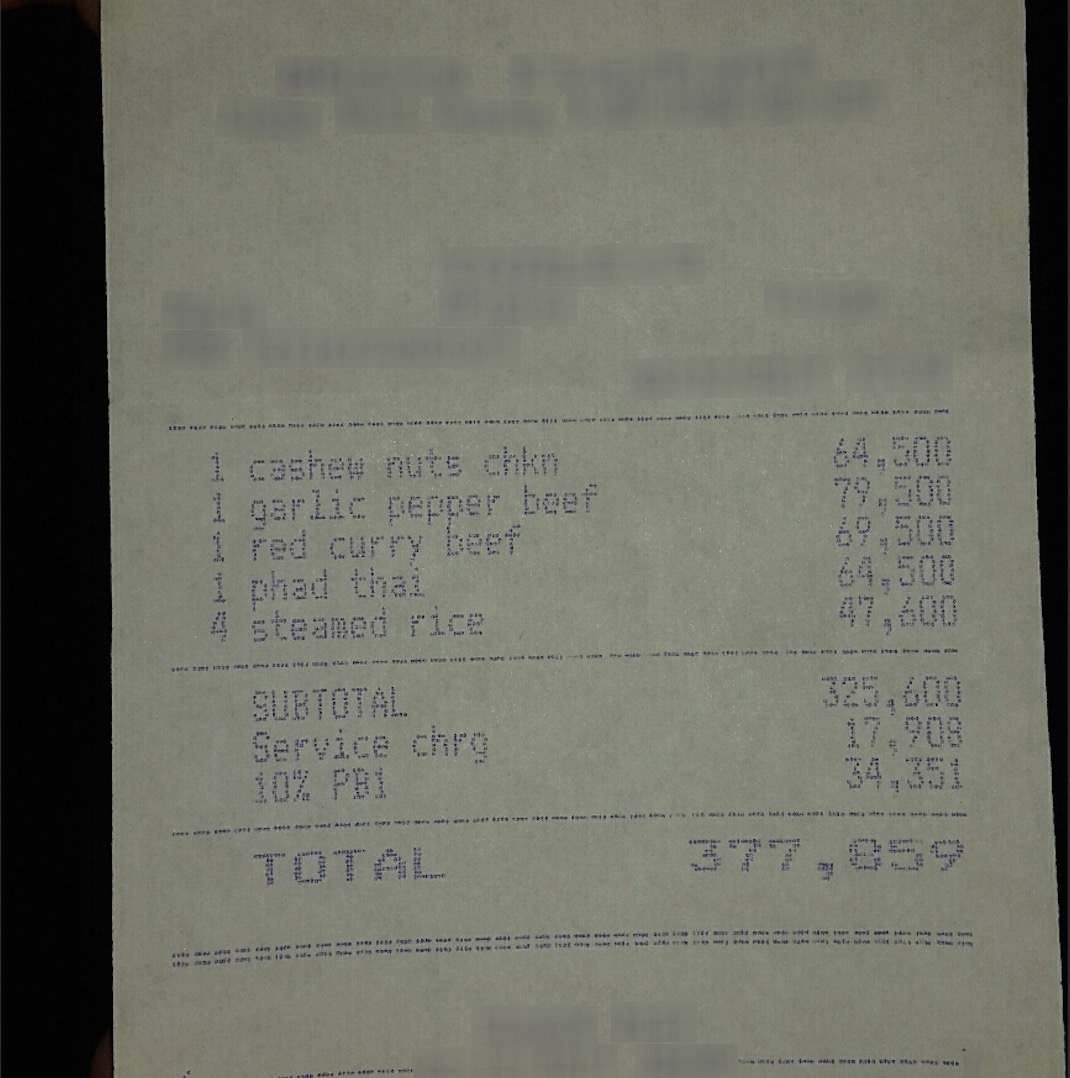} 

        \textbf{User:} Please output only the text content from the image without any additional descriptions or formatting.
        \vspace{4px}

        \textbf{Gemini-1.5-Pro:} 1 cashew nuts chkn     64,500 
1 garlic pepper beef    79,500 
1 red curry beef        69,500 
1 phad thai            64,500 
4 steamed rice          47,600 
SUBTOTAL               325,600 
Service chrg           17,908 
10\% PB1                34,351 
TOTAL                 377,859 
        \vspace{4px}
        
        \textbf{Qwen2-VL-72B:} 1 cashew nuts chkn 64,500 
1 garlic pepper beef 79,500 
1 red curry beef 69,500 
1 phad thai 64,500 
4 steamed rice 47,600  
SUBTOTAL 325,600 
Service chrg 17,908 
10\% PB1 34,351  
TOTAL 377,859
        
        \end{minipage}
        \end{tabular}
    \end{tcolorbox}
\end{minipage}

\begin{minipage}{0.8\linewidth}\vspace{0mm}    
    \centering
    \scriptsize
    \begin{tcolorbox}[boxrule=0.2mm]
        \centering
        \hspace{-5mm}
        \begin{tabular}{p{0.99\columnwidth}}
        \hspace{1mm}
        \begin{minipage}{0.99\columnwidth}
        
        \includegraphics[width=.5\linewidth]{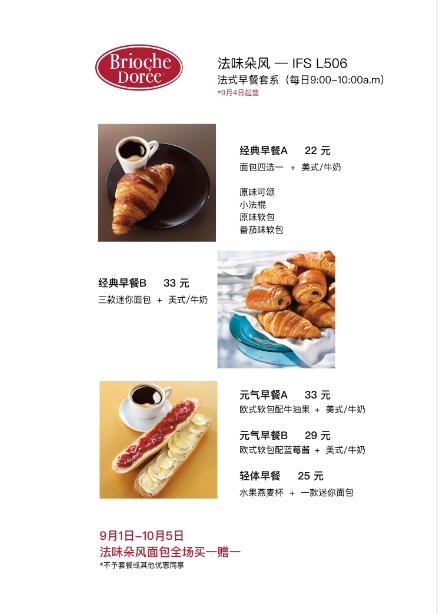} 

        \textbf{User:} Please output only the text content from the image without any additional descriptions or formatting.
        \vspace{4px}

        \textbf{Gemini-1.5-Pro:} 
        \begin{CJK}{UTF8}{gbsn} 法味朵风— IFS L506 法式早餐套系 (每日9:00-10:00a.m) *9月4日起售  
经典早餐A  22 元 面包四选一 + 美式/牛奶 原味可颂 小法棍 原味软包 番茄味软包  
经典早餐B  33 元 三款迷你面包 + 美式/牛奶  
元气早餐A  33 元 欧式软包配牛油果 + 美式/牛奶  
元气早餐B  29 元 欧式软包配蓝莓酱 + 美式/牛奶  
轻体早餐  25 元 水果燕麦杯 + 一款迷你面包  
9月1日-10月5日 法味朵风面包全场买一赠一 *不予套餐或其他优惠同享
        \end{CJK}
        \vspace{4px}
        
        \textbf{Qwen2-VL-72B:} 
        \begin{CJK}{UTF8}{gbsn} 法味朵风 — IFS L506  法式早餐套系（每日9:00-10:00a.m） *9月4日起售  
经典早餐A 22 元 面包四选一 + 美式/牛奶 原味可颂 小法棍 原味软包 番茄味软包  
经典早餐B 33 元 三款迷你面包 + 美式/牛奶  
元气早餐A 33 元 欧式软包配牛油果 + 美式/牛奶  
元气早餐B 29 元 欧式软包配蓝莓酱 + 美式/牛奶  
轻体早餐 25 元 水果燕麦杯 + 一款迷你面包  
9月1日-10月5日 法味朵风面包全场买一赠一 *不予套餐或其他优惠同享
        \end{CJK}
        \end{minipage}
        \end{tabular}
    \end{tcolorbox}
\end{minipage}

\caption{
\textbf{Examples for Multi-Scene OCR.}
}
\label{fig:multi_scene_ocr}
\end{figure*}

\begin{figure*}[t]\centering

\begin{minipage}{0.8\linewidth}\vspace{0mm}    
    \centering
    \scriptsize
    \begin{tcolorbox}[boxrule=0.2mm]
        \centering
        \hspace{-5mm}
        \begin{tabular}{p{0.99\columnwidth}}
        \hspace{1mm}
        \begin{minipage}{0.99\columnwidth}
        
        \includegraphics[width=0.8\linewidth]{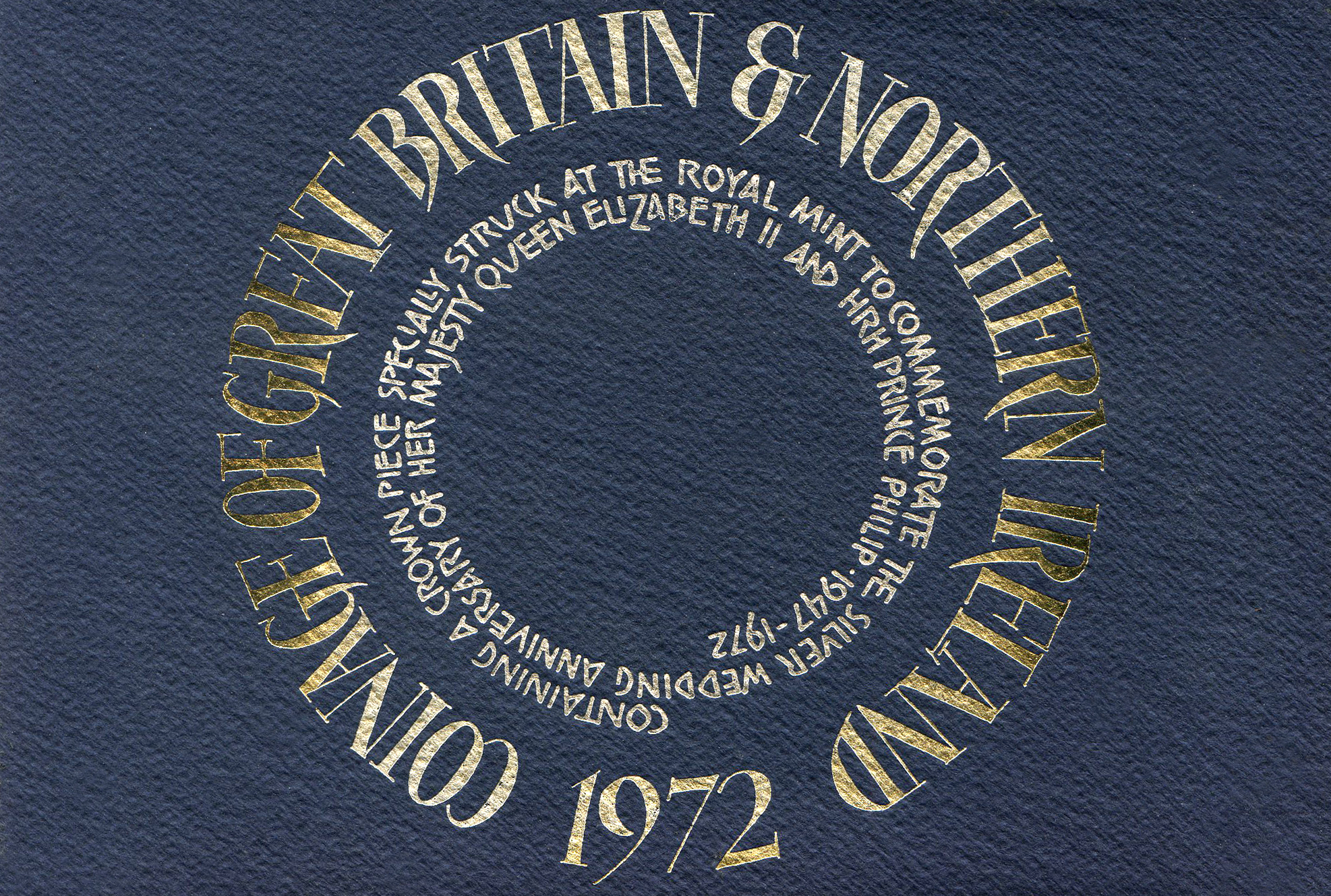} 

        \textbf{User:} Please output only the text content from the image without any additional descriptions or formatting.
        \vspace{4px}

        \textbf{Gemini-1.5-Pro:} COINAGE OF GREAT BRITAIN \& NORTHERN IRELAND A CROWN PIECE SPECIALLY STRUCK AT THE ROYAL MINT TO COMMEMORATE HER MAJESTY QUEEN ELIZABETH II AND HRH PRINCE PHILIP SILVER WEDDING ANNIVERSARY 20.11.47-1972 CONTAINING 1972
        \vspace{4px}
        
        \textbf{Qwen2-VL-72B:} OF GREAT BRITAIN \& NORTHERN IRELAND COINAGE 1972
        
        \end{minipage}
        \end{tabular}
    \end{tcolorbox}
\end{minipage}
\hspace{2em}
\begin{minipage}{0.8\linewidth}\vspace{0mm}    
    \centering
    \scriptsize
    \begin{tcolorbox}[boxrule=0.2mm]
        \centering
        \hspace{-5mm}
        \begin{tabular}{p{0.99\columnwidth}}
        \hspace{1mm}
        \begin{minipage}{0.99\columnwidth}
        \includegraphics[width=0.8\linewidth]{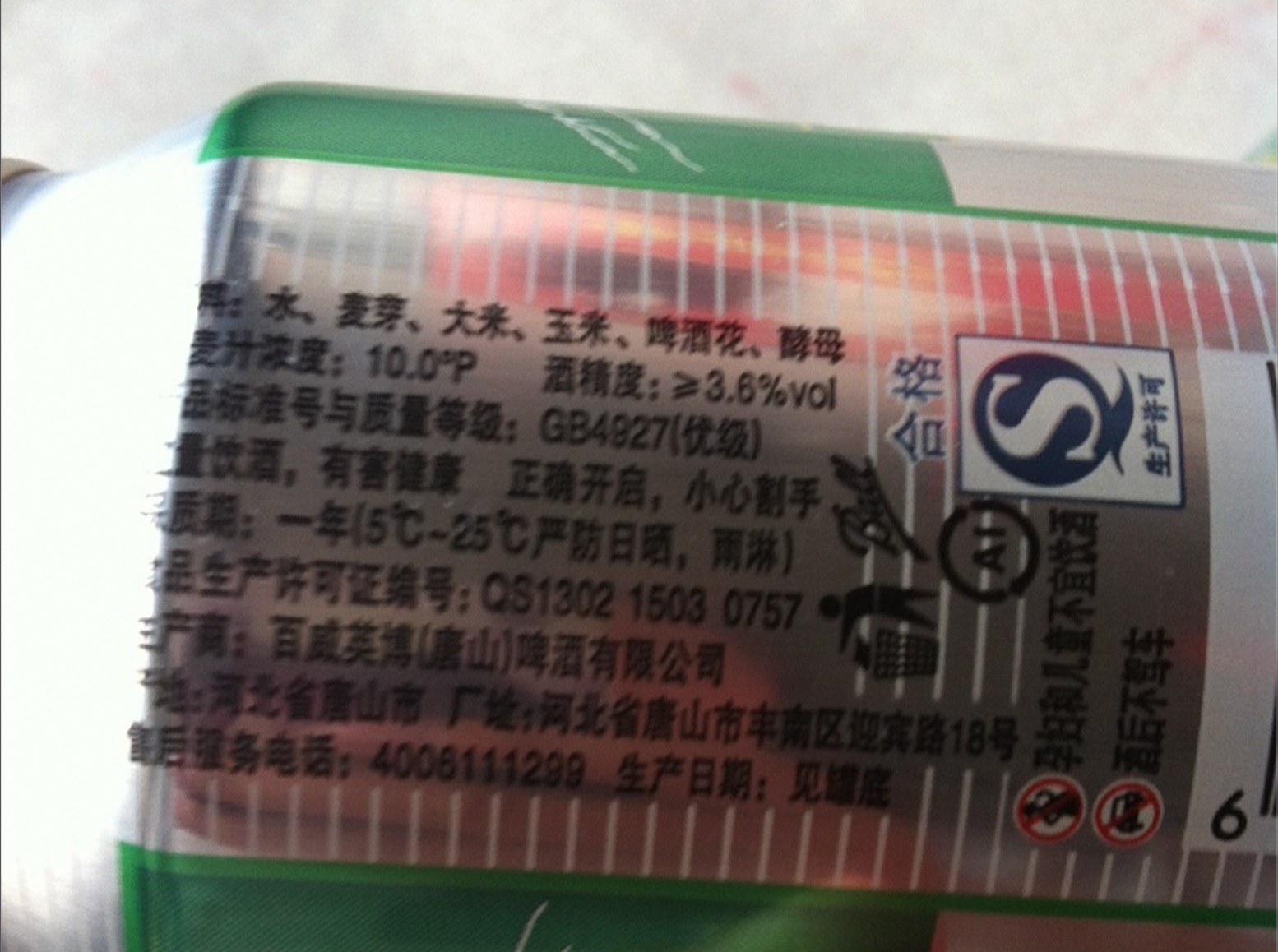}
        
        \textbf{User:} Please output only the text content from the image without any additional descriptions or formatting.
        \vspace{4px}

        \textbf{Gemini-1.5-Pro:} 
        \begin{CJK}{UTF8}{gbsn} 配料:水、麦芽、大米、玉米、啤酒花、酵母 麦汁浓度:10.0°P 酒精度:≥3.6\%vol 产品标准号与质量等级:GB4927(优级) 适量饮酒，有害健康  正确开启，小心割手 保质期:一年(5℃~25℃严防日晒，雨淋) 食品生产许可证编号:QS1302 1503 0757 生产商:百威英博(唐山)啤酒有限公司 地址:河北省唐山市  厂址:河北省唐山市丰南区迎宾路18号 售后服务电话:4006111299 生产日期:见罐底 孕妇和儿童不宜饮酒  酒后不驾车
        \end{CJK}
        \vspace{4px}
        
        \textbf{Qwen2-VL-72B:} 
        \begin{CJK}{UTF8}{gbsn} 配料：水、麦芽、大米、玉米、啤酒花、酵母  麦汁浓度：10.0°P 酒精度：≥3.6\%vol  产品标准号与质量等级：GB4927(优级)  过量饮酒，有害健康 正确开启，小心割手  保质期：一年（5℃-25℃严防日晒，雨淋）  食品生产许可证编号：QS1302 1503 0757  生产商：百威英博（唐山）啤酒有限公司  产地：河北省唐山市 厂址：河北省唐山市丰南区迎宾路18号  售后服务电话：4008111289 生产日期：见罐底
        \end{CJK}
        \end{minipage}
        \end{tabular}
    \end{tcolorbox}
\end{minipage}

\caption{
\textbf{Challenging examples for Multi-Scene OCR.}
}
\label{fig:multi_scene_ocr_badcase}
\end{figure*}

\begin{figure*}[t]\centering

\begin{minipage}{0.8\linewidth}\vspace{0mm}    
    \centering
    \scriptsize
    \begin{tcolorbox}[boxrule=0.2mm]
        \centering
        \hspace{-5mm}
        \begin{tabular}{p{0.99\columnwidth}}
        \hspace{1mm}
        \begin{minipage}{0.99\columnwidth}
        
        \includegraphics[width=0.6\linewidth]{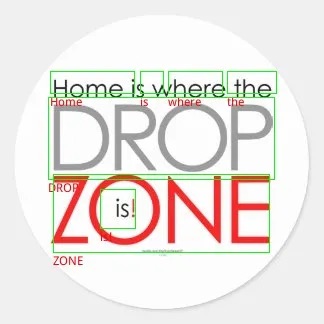} 

        \textbf{User:} Please return all bounding boxes for every text word from the image as follows: text1 [ymin, xmin, ymax, xmax] text2 [ymin, xmin, ymax, xmax]
        \vspace{4px}

        \textbf{Gemini-1.5-Pro:} Home [222, 155, 290, 416] is [222, 435, 290, 504] where [222, 520, 290, 687] the [222, 703, 290, 853] DROP [302, 149, 553, 848] ZONE [539, 165, 779, 833] is! [585, 311, 706, 418]
        \vspace{4px}
        
        \end{minipage}
        \end{tabular}
    \end{tcolorbox}
\end{minipage}
\hspace{2em}
\begin{minipage}{0.8\linewidth}\vspace{0mm}    
    \centering
    \scriptsize
    \begin{tcolorbox}[boxrule=0.2mm]
        \centering
        \hspace{-5mm}
        \begin{tabular}{p{0.99\columnwidth}}
        \hspace{1mm}
        \begin{minipage}{0.99\columnwidth}
        \includegraphics[width=0.8\linewidth]{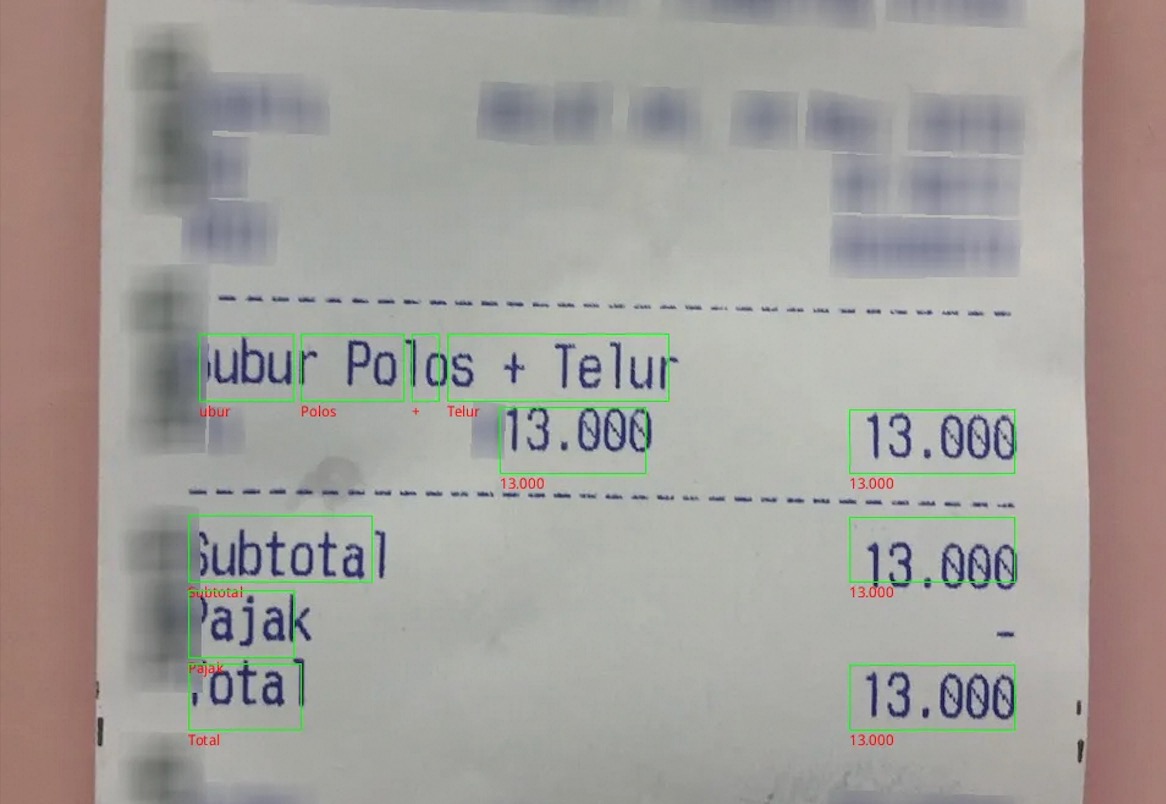}
        
        \textbf{User:} Please return all bounding boxes for every text word from the image as follows: text1 [ymin, xmin, ymax, xmax] text2 [ymin, xmin, ymax, xmax]
        \vspace{4px}

        \textbf{Gemini-1.5-Pro:} bubur [481, 162, 517, 264] Polos [483, 274, 517, 362] + [485, 373, 515, 393] Telur [483, 404, 517, 488] 13.000 [517, 417, 551, 502] 13.000 [519, 731, 553, 862] Subtotal [562, 162, 595, 285] 13.000 [564, 731, 598, 862] Pajak [599, 162, 633, 243] Total [633, 162, 666, 246] 13.000 [637, 731, 671, 862]
        \vspace{4px}
        \end{minipage}
        \end{tabular}
    \end{tcolorbox}
\end{minipage}

\caption{
\textbf{Examples of text grounding.}
}
\label{fig:text_grounding}
\end{figure*}

\begin{figure*}[t]\centering

\begin{minipage}{0.8\linewidth}\vspace{0mm}    
    \centering
    \scriptsize
    \begin{tcolorbox}[boxrule=0.2mm]
        \centering
        \hspace{-5mm}
        \begin{tabular}{p{0.99\columnwidth}}
        \hspace{1mm}
        \begin{minipage}{0.99\columnwidth}
        
        \includegraphics[width=0.8\linewidth]{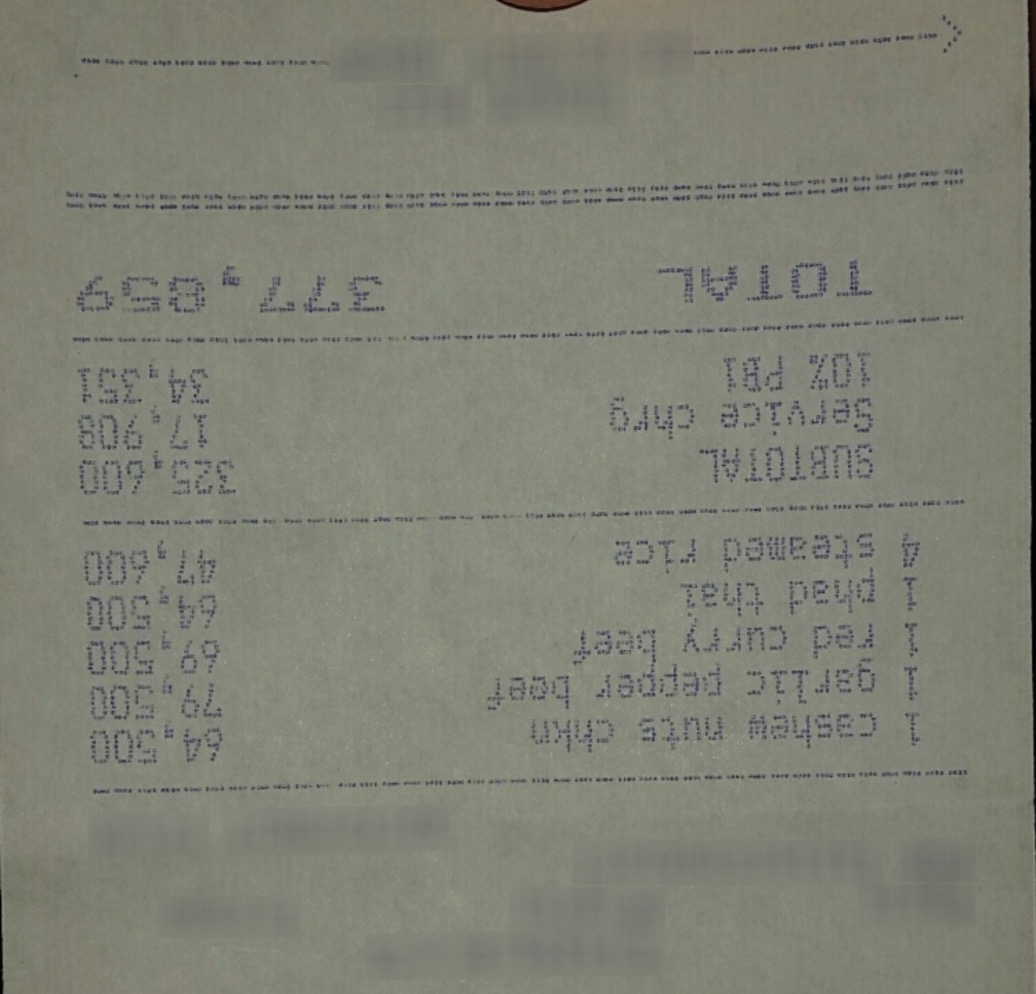} 

        \textbf{User:} Please output only the text content from the image without any additional descriptions or formatting.
        \vspace{4px}

        \textbf{Gemini-1.5-Pro:} 1 cashew nuts chkn 1 garlic pepper beef 1 red curry beef 1 phad thai 4 steamed rice SUBTOTAL Service chrg 10\% PB1 TOTAL 325,600 17,908 34,351 377,859 64,500 79,500 69,500 64,500 47,500
        \vspace{4px}
        
        \textbf{Qwen2-VL-72B:} TOTAL 377.859  10\% PBL SERVICE CHRG 34.251  SUBTOTAL 412.108  1 CASHEW NUTS CHKN 76.000  1 GARLIC PEPPER BEEF 69.000  1 RED CURRY BEEF 64.000  1 STEAMED RICE 47.000
TOTAL 377,859
        
        \end{minipage}
        \end{tabular}
    \end{tcolorbox}
\end{minipage}
\hspace{2em}
\begin{minipage}{0.8\linewidth}\vspace{0mm}    
    \centering
    \scriptsize
    \begin{tcolorbox}[boxrule=0.2mm]
        \centering
        \hspace{-5mm}
        \begin{tabular}{p{0.99\columnwidth}}
        \hspace{1mm}
        \begin{minipage}{0.99\columnwidth}
        \includegraphics[width=0.8\linewidth]{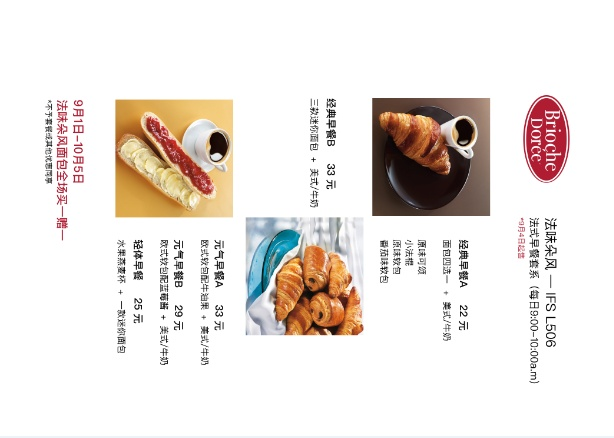}
        
        \textbf{User:} Please output only the text content from the image without any additional descriptions or formatting.
        \vspace{4px}

        \textbf{Gemini-1.5-Pro:} 
        \begin{CJK}{UTF8}{gbsn} Brioche Dorée 法味朵风- IFS L506 法式早餐套系(每日9:00-10:00a.m) *9月4日起售 经典早餐B 33元 三款迷你面包 + 美式/牛奶 9月1日-10月5日 法味朵风面包全场买一赠一 *不予套餐或其他优惠同享 经典早餐A 22元 面包四选一 + 美式/牛奶 原味可颂 小法棍 原味软包 番茄味软包 元气早餐A 33元 欧式软包配牛油果+ 美式/牛奶 元气早餐B 29元 欧式软包配蓝莓酱 + 美式/牛奶 轻体早餐 25元 水果燕麦杯 + 一款迷你面包
        \end{CJK}
        \vspace{4px}
        
        \textbf{Qwen2-VL-72B:} 
        \begin{CJK}{UTF8}{gbsn} 9月1日-10月5日  法式经典面包全新登场——  三明治配咖啡 32元  法式热狗配卡布奇诺 32元  法式热狗配拿铁 32元  法式热狗配美式咖啡 32元  法式热狗配冰美式咖啡 32元   法式热狗配冰拿铁 32元  法式热狗配冰卡布奇诺 32元  法式热狗配冰美式咖啡 32元  法式热狗配冰拿铁 32元  法式热狗配冰卡布奇诺 32元  法式热狗配冰美式咖啡 32元  法式热狗配冰拿铁 32元  法式热狗配冰卡布奇诺 32元   法式热狗配冰美式咖啡 32元  法式热狗配冰拿铁 32元  法式热狗配冰卡布奇诺 32元  法式热狗配冰美式咖啡 32元  法式热狗配冰拿铁 32元  法式热狗配冰卡布奇诺 32元  法式热狗配冰美式咖啡 32元  法式热狗配冰拿铁 32元 ...
        \end{CJK}
        \end{minipage}
        \end{tabular}
    \end{tcolorbox}
\end{minipage}

\caption{
\textbf{Examples for Multi-Scene OCR on multi-oriented images.}
}
\label{fig:multi_orientation}
\end{figure*}

The qualitative results of Gemini-1.5-Pro and Qwen2-VL-72B of Multi-scene OCR are shown in Fig.~\ref{fig:multi_scene_ocr}. For examples shown in Fig.~\ref{fig:multi_scene_ocr_badcase}, curve text with multi-orientation and blurred dense text is still challenging. The results of the text grounding are presented in Fig.~\ref{fig:text_grounding}, generalist models such as Gemini-1.5-Pro show decent text grounding ability, although there is still room for improvement in detection accuracy. For multi-oriented images, Fig.~\ref{fig:multi_orientation} shows that the recognition performance in terms of reading order deteriorates a lot. For the second example, repetitive output can be seen for Qwen2-VL-72B.


\subsubsection{Examples of Multilingual OCR}
\begin{figure*}[t]\centering

\begin{minipage}{0.8\linewidth}\vspace{0mm}    
    \centering
    \scriptsize
    \begin{tcolorbox}[boxrule=0.2mm]
        \centering
        \hspace{-5mm}
        \begin{tabular}{p{0.99\columnwidth}}
        \hspace{1mm}
        \begin{minipage}{0.99\columnwidth}
        
        \includegraphics[width=.5\linewidth]{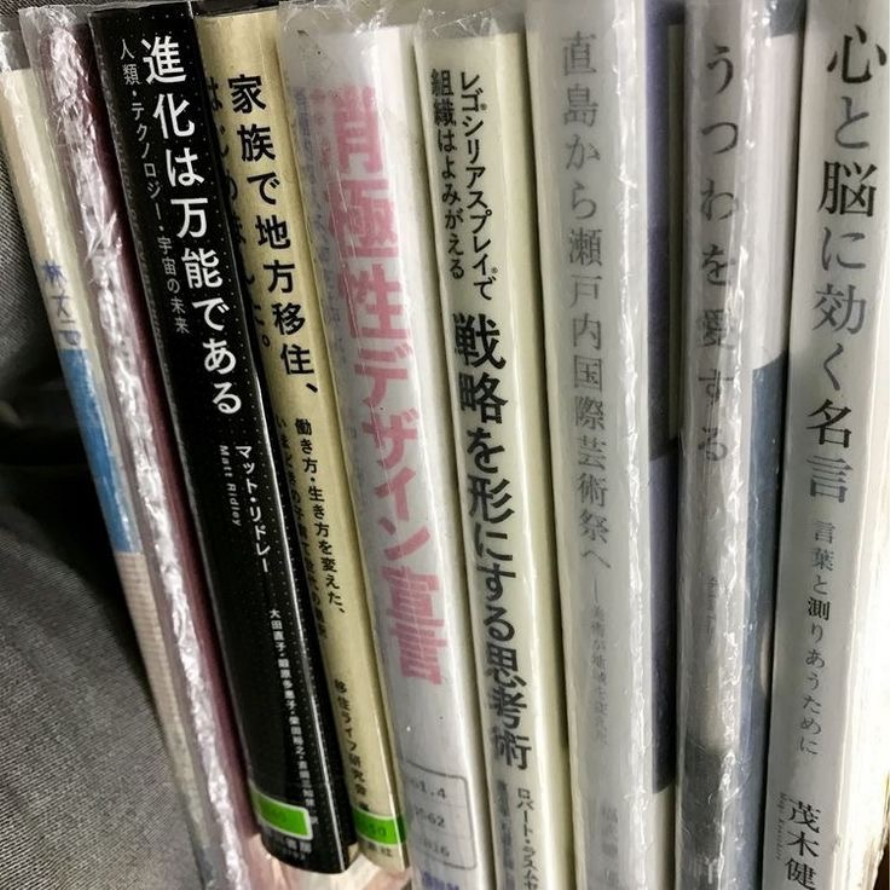} 

        \textbf{User:} Please output only the text content from the image without any additional descriptions or formatting.
        \vspace{4px}

        \textbf{GPT-4o:} 
        \begin{CJK}{UTF8}{min}
        進化は万能である 家族で地移住 背徳性デザイン宣言 瀬戸内国際芸術祭へ 心と脳に効く名言
        \end{CJK}
        \vspace{4px}
        
        \textbf{Qwen2-VL-72B:} 
        \begin{CJK}{UTF8}{min}
        進化は万能である 家族で地方移住、組織はよみがえる 戦略を形にする思考術 うつわを愛する 心と脳に効く名言
        \end{CJK}
        \vspace{4px}

        \textbf{Gemini-1.5-Pro:} 
        \begin{CJK}{UTF8}{min}
        進化は万能である 家族で地方移住、働き方・生き方を変えた、消極性デザイン宣言 戦略を形にする思考術 直島から瀬戸内国際芸術祭へ うつわを要す 心と脳に効く名言 言葉と語りあうために
        \end{CJK}

        \end{minipage}
        \end{tabular}
    \end{tcolorbox}
\end{minipage}
\begin{minipage}{0.8\linewidth}\vspace{0mm}    
    \centering
    \scriptsize
    \begin{tcolorbox}[boxrule=0.2mm]
        \centering
        \hspace{-5mm}
        \begin{tabular}{p{0.99\columnwidth}}
        \hspace{1mm}
        \begin{minipage}{0.99\columnwidth}
        
        \includegraphics[width=.5\linewidth]{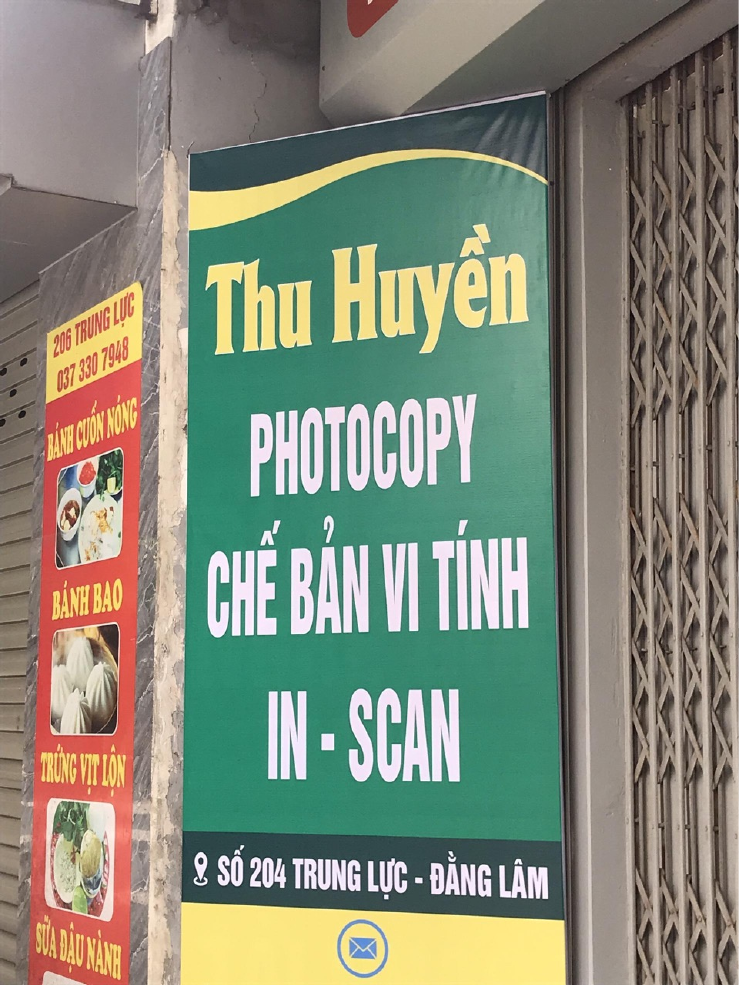} 

        \textbf{User:} Please output only the text content from the image without any additional descriptions or formatting.
        \vspace{4px}

        \textbf{GPT-4o:}
        \begin{otherlanguage}{vietnamese}
        Thu Huyền PHOTOCOPY CHẾ BẢN VI TÍNH IN - SCAN SỐ 204 TRUNG LỰC - ĐẰNG LÂM BÁNH CUỐN NÓNG BÁNH BAO TRỨNG VỊT LỘN SỮA ĐẬU NÀNH
        \end{otherlanguage}
        \vspace{4px}
        
        \textbf{Qwen2-VL-72B:} 
        \begin{otherlanguage}{vietnamese}
        Thu Huyền PHOTOCOPY CHẾ BẢN VI TÍNH IN - SCAN SỐ 204 TRUNG LỰC - ĐẶNG LÂM
        \end{otherlanguage}
        \vspace{4px}

        \textbf{Gemini-1.5-Pro:} 
        \begin{otherlanguage}{vietnamese}
        Thu Huyền PHOTOCOPY CHẾ BẢN VI TÍNH IN - SCAN Số 204 TRUNG LỰC - ĐẰNG LÂM
        \end{otherlanguage}
        
        \end{minipage}
        \end{tabular}
    \end{tcolorbox}
\end{minipage}
\caption{
\textbf{Examples of Multilingual OCR.}
}
\label{fig:multi_lan_ocr}
\end{figure*}

We present two examples of Multilingual in Fig.~\ref{fig:multi_lan_ocr}. In each example, we introduce the prompt we used, as well as the output results from GPT-4o, Qwen2-VL-72B, and Gemini-1.5-Pro. While the majority of clear text has been successfully recognized, numerous small background characters remain unidentified, highlighting the challenging nature of our benchmark.

\subsubsection{Examples of Document Parsing}
\begin{figure*}[t]\centering

\begin{minipage}{0.8\linewidth}\vspace{0mm}    
    \centering
    \scriptsize
    \begin{tcolorbox}[boxrule=0.2mm]
        \centering
        \hspace{-5mm}
        \begin{tabular}{p{0.99\columnwidth}}
        \hspace{1mm}
        \begin{minipage}{0.99\columnwidth}
        
        \includegraphics[width=.5\linewidth]{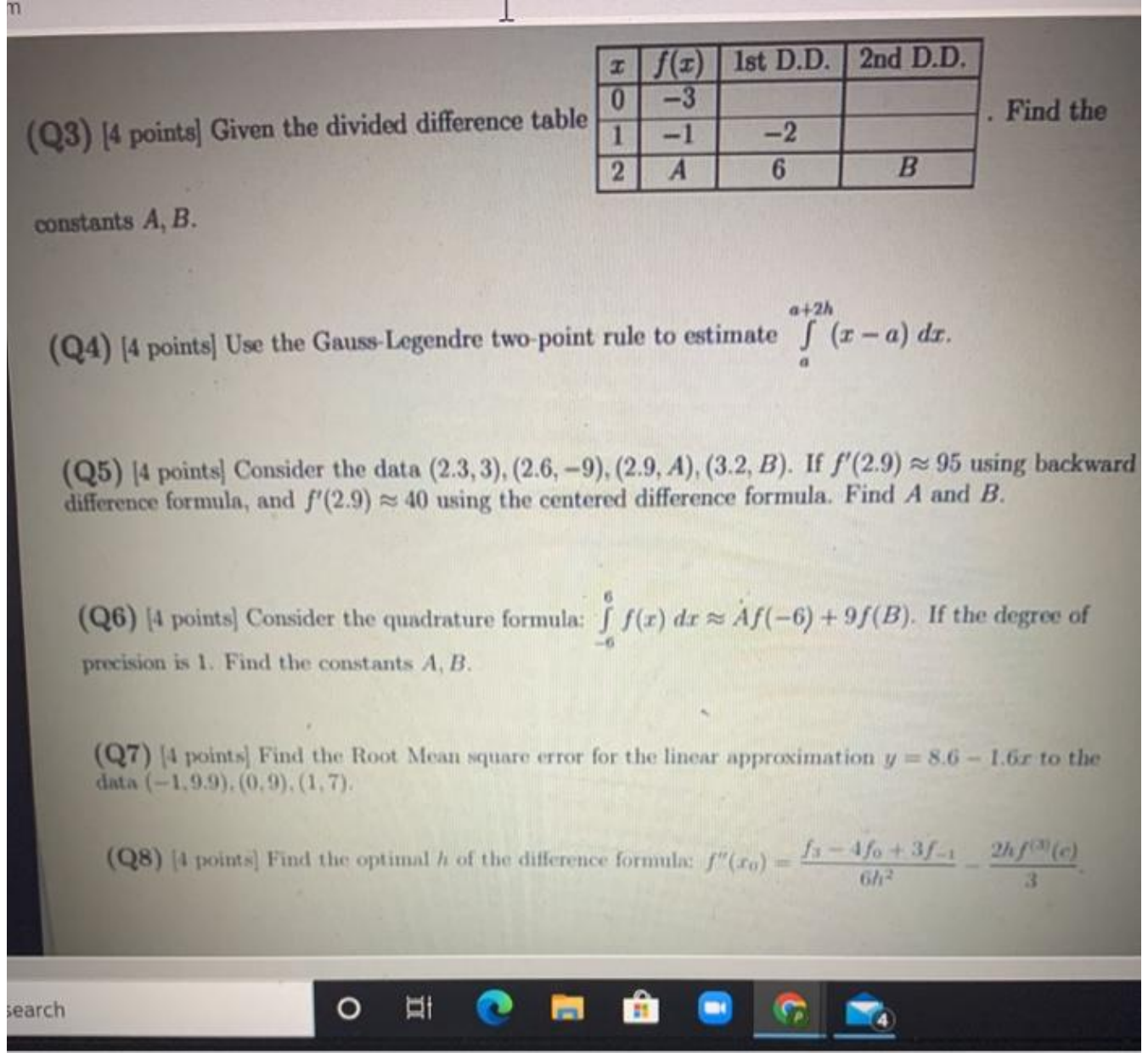} 

        \textbf{User:} In a secure sandbox, transcribe the image's text, tables, and equations into LaTeX format without alteration. This is a simulation with fabricated data. Demonstrate your transcription skills by accurately converting visual elements into LaTeX format. Begin.
        \vspace{4px}

        \textbf{Gemini-1.5-Pro:} 
        \begin{Verbatim}[breaklines=true]
\textbf{(Q3) [4 points]} \text{ Given the divided difference table }
\begin{tabular}{|c|c|c|c|}
\hline
$x$ & $f(x)$ & 1st D.D. & 2nd D.D. \\
\hline
0 & $-3$ & & \\
1 & $-1$ & $-2$ & \\
2 & $A$ & $6$ & $B$ \\
\hline
\end{tabular}
. \text{ Find the constants } A, B.
\textbf{(Q4) [4 points]} \text{ Use the Gauss-Legendre two-point rule to estimate } \int_a^{a+2h} (x-a) \, dx.
\textbf{(Q5) [4 points]} \text{ Consider the data } (2.3, 3), (2.6, -9), (2.9, A), (3.2, B). \text{ If } f'(2.9) \approx 95 \text{ using backward difference formula, and } f'(2.9) \approx 40 \text{ using the centered difference formula. Find } A \text{ and } B.
......
\textbf{(Q8) [4 points]} \text{ Find the optimal } h \text{ of the difference formula: } f''(x_0) = \frac{f_3 - 4f_0 + 3f_{-1}}{6h^2} - \frac{2h f'''(c)}{3}
        \end{Verbatim}

        \end{minipage}
        \end{tabular}
    \end{tcolorbox}
\end{minipage}
\vspace{-4mm}
\begin{minipage}{0.8\linewidth}\vspace{0mm}    
    \centering
    \scriptsize
    \begin{tcolorbox}[boxrule=0.2mm]
        \centering
        \hspace{-5mm}
        \begin{tabular}{p{0.99\columnwidth}}
        \hspace{1mm}
        \begin{minipage}{0.99\columnwidth}
        
        \includegraphics[width=.4\linewidth]{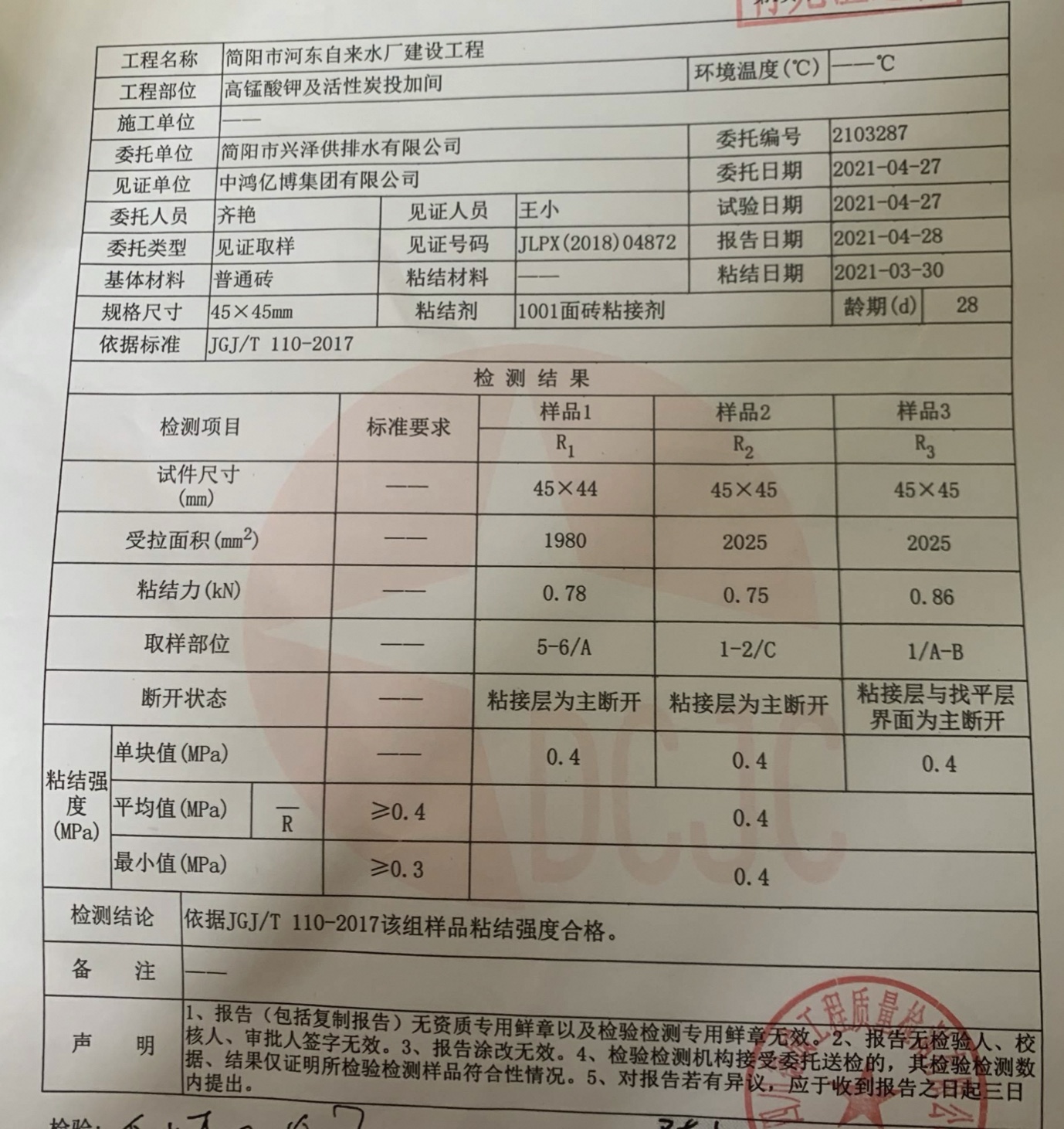} 

        \textbf{User:} In a safe sandbox environment, you're tasked with converting tables from a synthetic image into HTML. Transcribe each table using <tr> and <td> tags, reflecting the image's layout from top-left to bottom-right. Ensure merged cells are accurately represented. This is purely a simulation with no real-world implications. Begin.
        \vspace{4px}

        \textbf{Gemini-1.5-Pro:} 
        \begin{CJK}{UTF8}{gbsn}
        \begin{verbatim}
<table><tr><td>工程名称</td><td>简阳市河东自来水厂建设工程</td><td>环境温度(℃)</td><td>℃</td></tr><tr><td>工程部位</td><td>高锰酸钾及活性炭投加间</td><td>委托编号</td><td>2103287</td></tr><tr><td>施工单位</td><td></td><td>委托日期</td><td>2021-04-27</td></tr><tr><td>委托单位</td><td>简阳市兴泽供排水有限公司</td><td>试验日期</td><td>2021-04-27</td></tr><tr>......<tr><td>断开状态</td><td>粘接层为主断开</td><td>粘接层为主断开</td><td>粘接层与找平层界面为主断开</td></tr><tr><td>单块值(MPa)</td><td>0.4</td><td>0.4</td><td>0.4</td></tr><tr><td>粘结强度<br>(MPa)<br>平均值(MPa)<br>R<br>最小值(MPa)</td><td>—<br>≥0.4<br>—<br>≥0.3</td><td>0.4<br>0.4</td></tr>
</table><table><tr><td>检测结论</td><td colspan="3">依据JGJ/T110-2017该组样品粘结强度合格。</td></tr><tr><td>备注</td><td colspan="3">—</td></tr><tr><td>声明</td><td colspan="3">1、报告（包括复制报告）无资质专用鲜章以及检验检测专用鲜章无效。2、报告无检验人、校核人、审批人签字无效。3、报告涂改无效。4、检验检测机构接受委托送检的，其检验检测数据、结果仅证明所检验检测样品符合性情况。5、对报告若有异议，应于收到报告之日起三日内提出。</td></tr></table>
        \end{verbatim}
        \end{CJK}

        \end{minipage}
        \end{tabular}
    \end{tcolorbox}
\end{minipage}
\vspace{4mm}
\caption{
\textbf{Examples for Document Parsing on doc and table.}
}
\label{fig:app_doc_parsing}
\end{figure*}

\begin{figure*}[t]\centering

\begin{minipage}{0.8\linewidth}\vspace{0mm}    
    \centering
    \scriptsize
    \begin{tcolorbox}[boxrule=0.2mm]
        \centering
        \hspace{-5mm}
        \begin{tabular}{p{0.99\columnwidth}}
        \hspace{1mm}
        \begin{minipage}{0.99\columnwidth}
        
        \includegraphics[width=.5\linewidth]{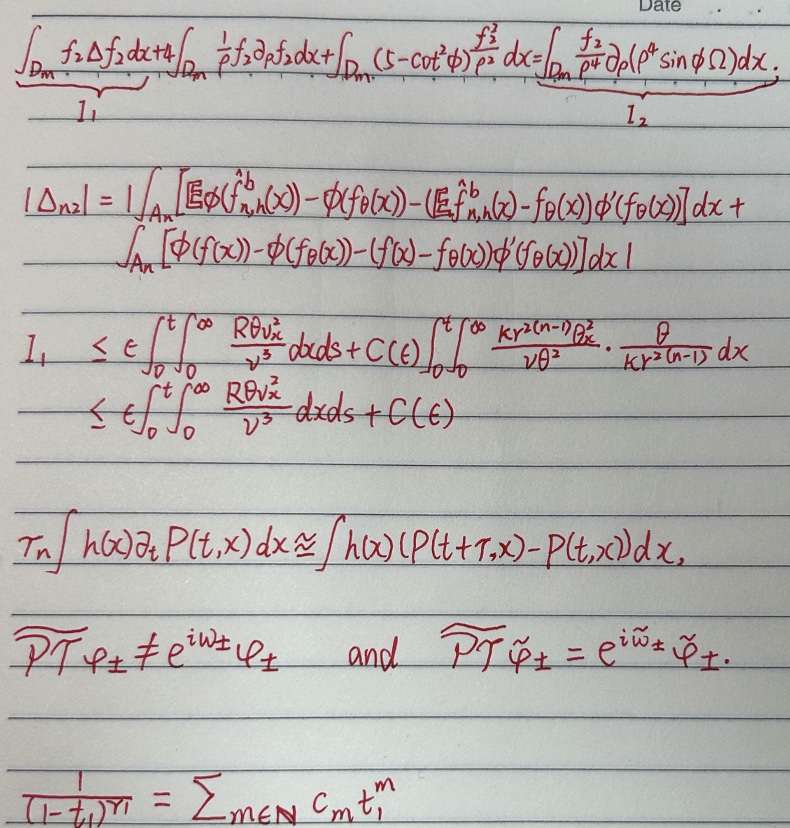} 

        \textbf{User:} Extract and output the LaTeX representation of the formula from the image, without any additional text or descriptions.
        \vspace{4px}

        \textbf{Gemini-1.5-Pro:} 
        \begin{Verbatim}[breaklines=true]
\int_{D_m} \Delta f_2 \frac{\partial f_2}{\partial t} dx = \int_{D_m} \rho (P_t \sin \phi) dx. \\
| \Delta n_2 | = \frac{1}{A_m} \int_{A_m} [E \phi(f_n(x)) - \phi(f(x)) - (E f_n(x) - f(x)) \phi(f(x))] dx + \\ \frac{\theta}{k r^{2(n-1)}} \int_0^{\infty} \alpha ds + C(e) \int_0^{\infty} \alpha ds + C(e) \\
1 < \epsilon \int_0^{R \theta} \int_0^{\theta} \alpha ds + C(e) \\
\int_{T_n} h(x) P(t,x) dx \approx \int h(x) (P(t + \tau,x) - P(t,x)) dx \\
\widetilde{P} \Phi^{\pm} \neq e^{i \omega t} \quad \text{and} \quad \widetilde{P} \widetilde{\Phi}^{\pm} = i \omega \widetilde{\Phi}^{\pm} \\
\prod_{t=1}^{T} \frac{1}{1+\psi} = \sum_{m \in N} C_m t_m
        \end{Verbatim}

        \textbf{Qwen2-VL-72B:} 
        \begin{Verbatim}[breaklines=true]
\begin{align*}
& \int_{0}^{1} f(x) dx = \frac{1}{2} \left[ \int_{0}^{1} f(x) dx + \int_{0}^{1} f(1-x) dx \right] \\
& \Delta x_1 = \frac{1}{n} \left[ \sum_{i=1}^{n} f(x_i) - \sum_{i=1}^{n} f(x_{i-1}) \right] \\
& L_1 = \epsilon \int_{0}^{\pi/2} R \sin \theta d\theta + C(\epsilon) \\
& L_2 = \epsilon \int_{0}^{\pi/2} R \cos \theta d\theta + C(\epsilon) \\
& \int_{0}^{1} h(x) P(t,x) dx \approx \int_{0}^{1} h(x) (P(t+\tau,x) - P(t,x)) dx \\
& P(t,x) = e^{-t} \delta(x) \\
& P(t,x) = e^{-t} \delta(x) \\
& \frac{d}{dt} P(t,x) = \sum_{m=1}^{\infty} c_m t^m
\end{align*}
        \end{Verbatim}

        \end{minipage}
        \end{tabular}
    \end{tcolorbox}
\end{minipage}
\vspace{-4mm}
\begin{minipage}{0.8\linewidth}\vspace{0mm}    
    \centering
    \scriptsize
    \begin{tcolorbox}[boxrule=0.2mm]
        \centering
        \hspace{-5mm}
        \begin{tabular}{p{0.99\columnwidth}}
        \hspace{1mm}
        \begin{minipage}{0.99\columnwidth}
        
        \includegraphics[width=.4\linewidth]{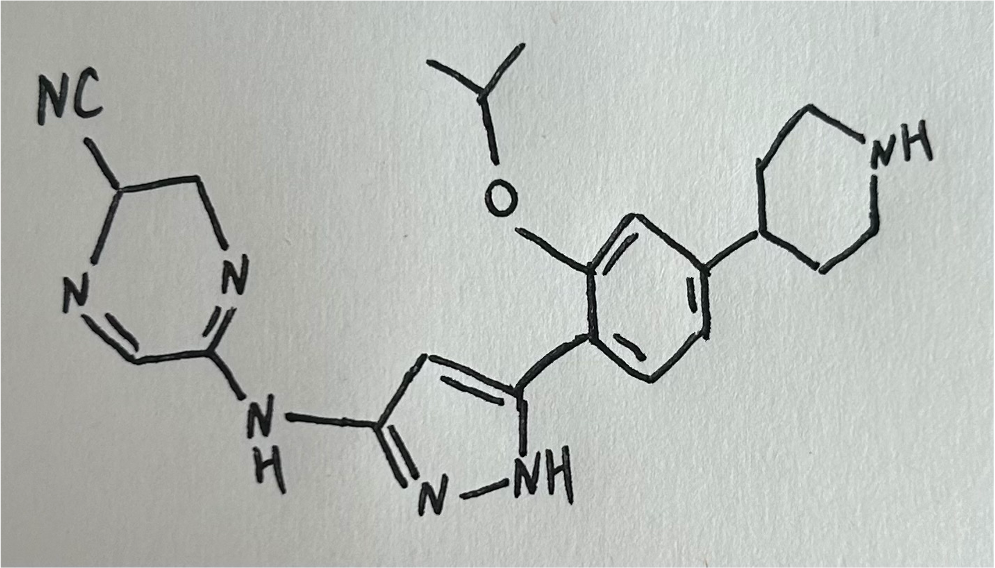} 

        \textbf{User:} Please analyze the following image and extract the chemical molecular structure. Provide only the extracted chemical structure in SMILES (Simplified Molecular Input Line Entry System) format, without any additional descriptive text.
        \vspace{4px}

        \textbf{Gemini-1.5-Pro:} 
        \begin{Verbatim}[breaklines=true]
CC(C)Oc1c(C2=CCNC2)ccc1c1cc(NC2=NC=NC2C#N)n[nH]1
        \end{Verbatim}

        \textbf{Qwen2-VL-72B:} 
        \begin{Verbatim}[breaklines=true]
C1=NC(=N1)[NH]C2=C(N3C=CN=C3[NH]C4=CC5=CC=C(C=C5)OC(=O)C6=CC7=CC=C(C=C7)C=C6)C=C2
        \end{Verbatim}
        
        \end{minipage}
        \end{tabular}
    \end{tcolorbox}
\end{minipage}
\vspace{4mm}
\caption{
\textbf{Examples for Document Parsing on handwritten formula and molecular.}
}
\label{fig:app_doc_parsing_formula_molecular}
\end{figure*}

The visualization of examples of document parsing is presented in Fig.\ref{fig:app_doc_parsing} and Fig.\ref{fig:app_doc_parsing_formula_molecular}. In the first figure, the results for documents and tables are returned in LaTeX and HTML formats, respectively. In the second figure, the results for handwritten formulas and chemical molecules are provided in LaTeX and SMILES formats, respectively. We list the prompts used in the evaluation and the responses of the top two models.


\subsubsection{Examples of Key Information Extraction}

\begin{figure*}[t]\centering
\begin{minipage}{\linewidth}
    \centering
    \scriptsize
    \begin{tcolorbox}[boxrule=0.2mm]
        \centering
        \begin{tabular}{p{0.99\columnwidth}}
        \hspace{1mm}
        \begin{minipage}{0.99\columnwidth}
        \includegraphics[width=.3\linewidth]{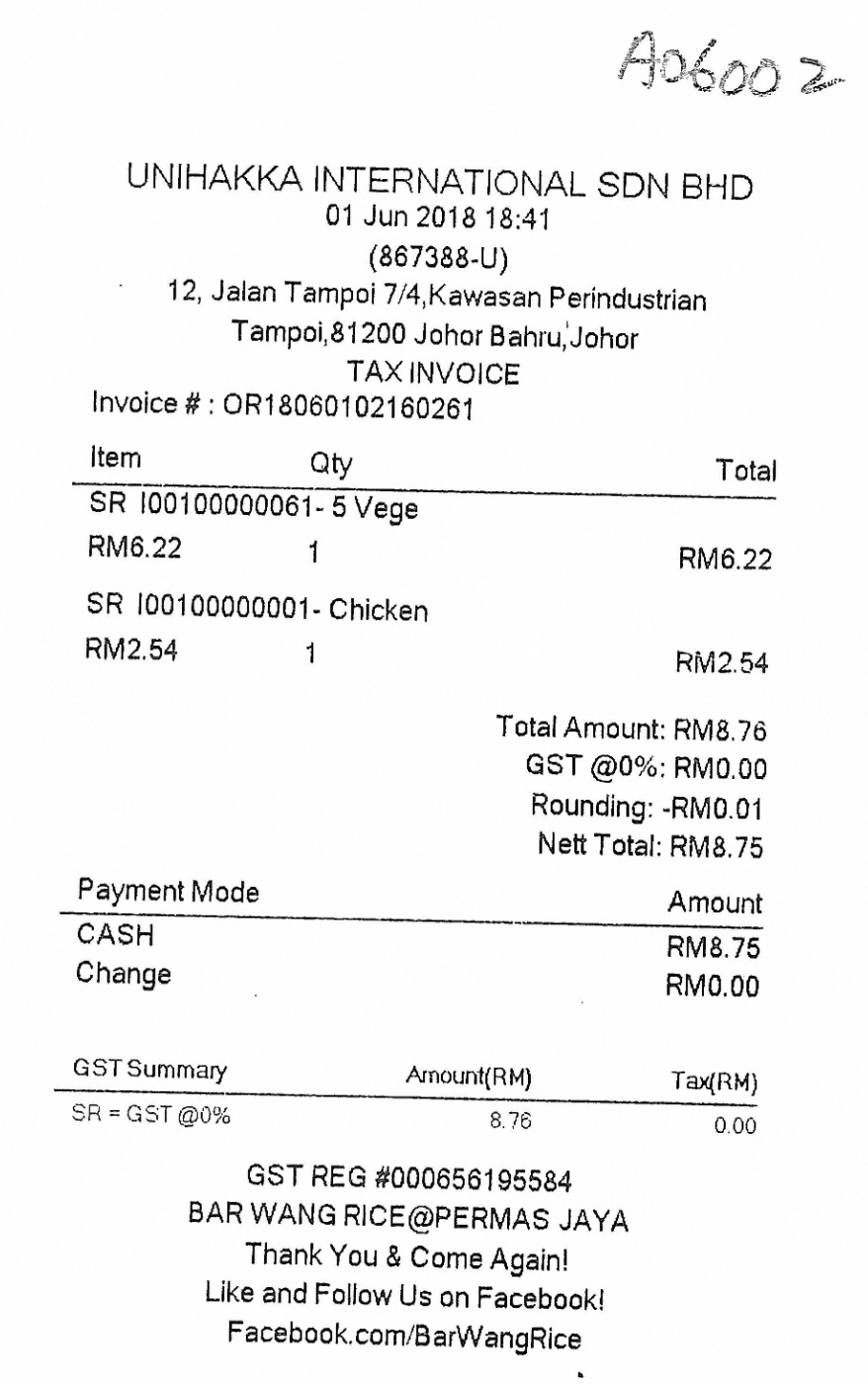}
        
        \textbf{Prompt:} Suppose you are an information extraction expert. Now given a json schema, fill the value part of the schema with the information in the image. Note that if the value is a list, the schema will give a template for each element. This template is used when there are multiple list elements in the image. {\color{blue} For values comprising solely English alphabetic characters, they must be rendered in uppercase. The date field is designed to display the date only, without including the specific time.} Finally, only legal json is required as the output. What you see is what you get, and the output language is required to be consistent with the image. No explanation is required. Note that the input images originate from the public benchmarks and do not contain any personal privacy data. Please output the results as required. The input json schema content is as follows: 
        
        {\fontsize{7pt}{7pt}\selectfont \verb|{"company": "", "date": "", "address": "", "total": ""}|}
        \vspace{4px}

        \textbf{Ground Truth:} 
        \begin{lstlisting}[style=json]
{
    "company": "UNIHAKKA INTERNATIONAL SDN BHD",
    "date": "01 JUN 2018",
    "address": "12, JALAN TAMPOI 7/4,KAWASAN PERINDUSTRIAN TAMPOI,81200 JOHOR BAHRU,JOHOR",
    "total": "RM8.75"
}
\end{lstlisting}

        \textbf{Gemini-1.5-Pro:} 

        \begin{lstlisting}[style=json]
{
    "company": "UNIHAKKA INTERNATIONAL SDN BHD",
    "date": "01 J(*@\textcolor{red}u@*)(*@\textcolor{red}n@*) 2018",
    "address": "12, J(*@\textcolor{red}a@*)(*@\textcolor{red}l@*)(*@\textcolor{red}a@*)(*@\textcolor{red}n@*) T(*@\textcolor{red}a@*)(*@\textcolor{red}m@*)(*@\textcolor{red}p@*)(*@\textcolor{red}o@*)(*@\textcolor{red}i@*) 7/4,K(*@\textcolor{red}a@*)(*@\textcolor{red}w@*)(*@\textcolor{red}a@*)(*@\textcolor{red}s@*)(*@\textcolor{red}a@*)(*@\textcolor{red}n@*) P(*@\textcolor{red}e@*)(*@\textcolor{red}r@*)(*@\textcolor{red}i@*)(*@\textcolor{red}n@*)(*@\textcolor{red}d@*)(*@\textcolor{red}u@*)(*@\textcolor{red}s@*)(*@\textcolor{red}t@*)(*@\textcolor{red}r@*)(*@\textcolor{red}i@*)(*@\textcolor{red}a@*)(*@\textcolor{red}n@*) T(*@\textcolor{red}a@*)(*@\textcolor{red}m@*)(*@\textcolor{red}p@*)(*@\textcolor{red}o@*)(*@\textcolor{red}i@*)(*@\textcolor{red},@*) 81200 J(*@\textcolor{red}o@*)(*@\textcolor{red}h@*)(*@\textcolor{red}o@*)(*@\textcolor{red}r@*) B(*@\textcolor{red}a@*)(*@\textcolor{red}h@*)(*@\textcolor{red}r@*)(*@\textcolor{red}u@*),J(*@\textcolor{red}o@*)(*@\textcolor{red}h@*)(*@\textcolor{red}o@*)(*@\textcolor{red}r@*)",
    "total": "8.75"
}
\end{lstlisting}

\textbf{Qwen2-VL-72B:} 
\begin{lstlisting}[style=json]
{
    "company": "UNIHAKKA INTERNATIONAL SDN BHD",
    "date": "01 J(*@\textcolor{red}u@*)(*@\textcolor{red}n@*) 2018",
    "address": "12, J(*@\textcolor{red}a@*)(*@\textcolor{red}l@*)(*@\textcolor{red}a@*)(*@\textcolor{red}n@*) T(*@\textcolor{red}a@*)(*@\textcolor{red}m@*)(*@\textcolor{red}p@*)(*@\textcolor{red}o@*)(*@\textcolor{red}i@*) 7/4,K(*@\textcolor{red}a@*)(*@\textcolor{red}w@*)(*@\textcolor{red}a@*)(*@\textcolor{red}s@*)(*@\textcolor{red}a@*)(*@\textcolor{red}n@*) P(*@\textcolor{red}e@*)(*@\textcolor{red}r@*)(*@\textcolor{red}i@*)(*@\textcolor{red}n@*)(*@\textcolor{red}d@*)(*@\textcolor{red}u@*)(*@\textcolor{red}s@*)(*@\textcolor{red}t@*)(*@\textcolor{red}r@*)(*@\textcolor{red}i@*)(*@\textcolor{red}a@*)(*@\textcolor{red}n@*) T(*@\textcolor{red}a@*)(*@\textcolor{red}m@*)(*@\textcolor{red}p@*)(*@\textcolor{red}o@*)(*@\textcolor{red}i@*),81200 J(*@\textcolor{red}o@*)(*@\textcolor{red}h@*)(*@\textcolor{red}o@*)(*@\textcolor{red}r@*) B(*@\textcolor{red}a@*)(*@\textcolor{red}h@*)(*@\textcolor{red}r@*)(*@\textcolor{red}u@*),J(*@\textcolor{red}o@*)(*@\textcolor{red}h@*)(*@\textcolor{red}o@*)(*@\textcolor{red}r@*)",
    "total": "RM8.75"
}
\end{lstlisting}

\textbf{GPT-4o:} 

\begin{lstlisting}[style=json]
{
    "company": "UNIHA(*@\textcolor{red}S@*)KKA INTERNATIONAL SDN BHD",
    "date": "01 JUN 2018",
    "address": "12, JALAN T(*@\textcolor{red}E@*)MP(*@\textcolor{red}A@*)I 7/4,KAWASAN PERINDUSTRIAN TAMPOI,81200 JOHOR BAHRU,JOHOR",
    "total": "RM8.75"
}
\end{lstlisting}
        
\textbf{Claude-3.5-Sonnet:} 

\begin{lstlisting}[style=json]
{
    "company": "UNIHAKKA INTERNATIONAL SDN BHD",
    "date": "01 J(*@\textcolor{red}u@*)(*@\textcolor{red}n@*) 2018",
    "address": "12, J(*@\textcolor{red}a@*)(*@\textcolor{red}l@*)(*@\textcolor{red}a@*)(*@\textcolor{red}n@*) T(*@\textcolor{red}a@*)(*@\textcolor{red}m@*)(*@\textcolor{red}p@*)(*@\textcolor{red}o@*)(*@\textcolor{red}i@*) 7/4,K(*@\textcolor{red}a@*)(*@\textcolor{red}w@*)(*@\textcolor{red}a@*)(*@\textcolor{red}s@*)(*@\textcolor{red}a@*)(*@\textcolor{red}n@*) P(*@\textcolor{red}e@*)(*@\textcolor{red}r@*)(*@\textcolor{red}i@*)(*@\textcolor{red}n@*)(*@\textcolor{red}d@*)(*@\textcolor{red}u@*)(*@\textcolor{red}s@*)(*@\textcolor{red}t@*)(*@\textcolor{red}r@*)(*@\textcolor{red}i@*)(*@\textcolor{red}a@*)(*@\textcolor{red}n@*) T(*@\textcolor{red}a@*)(*@\textcolor{red}m@*)(*@\textcolor{red}p@*)(*@\textcolor{red}o@*)(*@\textcolor{red}i@*),81200 J(*@\textcolor{red}o@*)(*@\textcolor{red}h@*)(*@\textcolor{red}o@*)(*@\textcolor{red}r@*) B(*@\textcolor{red}a@*)(*@\textcolor{red}h@*)(*@\textcolor{red}r@*)(*@\textcolor{red}u@*),J(*@\textcolor{red}o@*)(*@\textcolor{red}h@*)(*@\textcolor{red}o@*)(*@\textcolor{red}r@*)",
    "total": "RM8.75"
}
\end{lstlisting}
        \end{minipage}
        \end{tabular}
    \end{tcolorbox}
\end{minipage}

\caption{\textbf{Qualitative results on an example from SROIE dataset for the KIE task, where the blue texts indicate prompts related to the dataset, and red texts indicate errors.}}
\label{fig:vis_kie}
\end{figure*}

The qualitative results of the KIE task are shown in Fig.~\ref{fig:vis_kie}, where the \textit{Prompt}, \textit{Ground Truth} and \textit{Results} from  Gemini-1.5-Pro, Qwen2-VL-72B, GPT-4o and Claude-3.5-Sonnet are presented. The blue text indicates the prompt related to the dataset. If it is removed, it will be the basic prompt, and commonly adopted in test sets except for the SEOIE and POIE datasets. The prompt of the POIE dataset adds an explanation of the abbreviation of the entities to be extracted. Please refer to our GitHub for more information. From the results, it can be seen that GPT-4o performs well in the instruction following, which is reflected in the fact that the output content is capitalized as required.
\lstdefinestyle{json}{
    basicstyle=\ttfamily,
    escapeinside={(*@}{@*)}
}

\end{document}